\documentclass[10pt,twocolumn,letterpaper]{article}

\usepackage{cvpr}              

\definecolor{cvprblue}{rgb}{0.21,0.49,0.74}
\usepackage{graphicx}
\usepackage{amsmath}
\usepackage{mathtools}
\usepackage{amssymb}
\usepackage{amsfonts}
\usepackage{booktabs}
\usepackage[table,dvipsnames]{xcolor}
\usepackage{multirow}
\usepackage{subcaption}
\usepackage{caption}

\usepackage[pagebackref,breaklinks=true,colorlinks,citecolor=blue,urlcolor=blue,linkcolor=blue,bookmarks=false]{hyperref}

\def\paperID{10301} 
\def\confName{CVPR}
\def\confYear{2026}

\title{Does Hearing Help Seeing? \\ Investigating Audio–Video Joint Denoising for Video Generation}

\author{
    Jianzong Wu$^{1,2}$ \quad
    Hao Lian$^1$ \quad
    Dachao Hao$^1$ \quad
    Ye Tian$^1$ \quad
    Qingyu Shi$^1$ \quad \\
    Biaolong Chen$^2$ \quad
    Hao Jiang$^2$ \quad
    Yunhai Tong$^{1}$ \\
   {\normalsize $^1$ Peking University \quad $^2$ Alibaba Group} \\
    {\normalsize Project Page: \url{https://jianzongwu.github.io/projects/does-hearing-help-seeing/}} \\
    {\normalsize \textit{Email: jzwu@stu.pku.edu.cn}}
}

\begin{document}
\twocolumn[{%
\renewcommand\twocolumn[1][]{#1}%
\maketitle
\centering
\vspace{-5mm}
\includegraphics[width=1.0\linewidth]{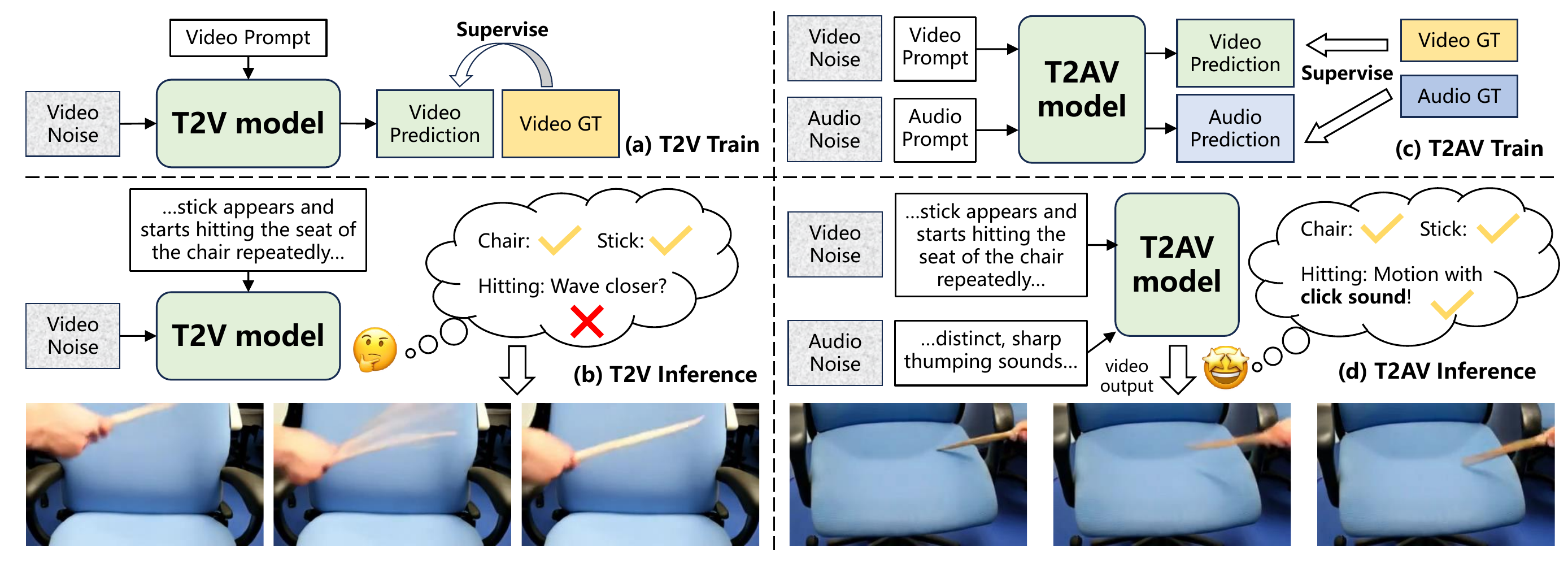}
\captionof{figure}{\textbf{Conceptual comparison between T2V and T2AV diagrams.} (a) T2V training denoises video latents with video-only supervision. 
(b) In inference, T2V may misinterpret motion because there is insufficient evidence linking visual appearances to world physics. 
(c) T2AV training jointly denoises audio and video latents with audio–video supervision. 
(d) In inference, T2AV produces physically plausible motion with synchronized audio. Audio helps video generation models understand the world.}
\label{fig:teaser}
\vspace{5mm}
}]

\begin{abstract}
Recent audio-video generative systems suggest that coupling modalities benefits not only audio–video synchrony but also the video modality itself. 
We pose a fundamental question: Does audio–video joint denoising training improve video generation, even when we only care about video quality? 
To study this, we introduce a parameter-efficient Audio–Video Full DiT (AVFullDiT) architecture that leverages pre-trained text-to-video (T2V) and text-to-audio (T2A) modules for joint denoising.
We train (i) a T2AV model with AVFullDiT and (ii) a T2V-only counterpart under identical settings. Our results provide the first systematic evidence that audio–video joint denoising can deliver more than synchrony. 
We observe consistent improvements on challenging subsets featuring large and object contact motions. 
We hypothesize that predicting audio acts as a privileged signal, encouraging the model to internalize causal relationships between visual events and their acoustic consequences (e.g., collision $\times$ impact sound), which in turn regularizes video dynamics. 
Our findings suggest that cross-modal co-training is a promising approach to developing stronger, more physically grounded world models.
Code and dataset will be made publicly available.
\end{abstract}
\vspace{-8mm}
\section{Introduction}

Recent closed-source audio–video (A/V) generators, such as Sora 2~\cite{Sora2} and Veo 3~\cite{Veo3}, demonstrate not only striking audio–video synchronization but also marked advances in video-only fidelity, particularly for large, fast motions and physically grounded dynamics that have challenged prior text-to-video (T2V) systems~\cite{zeroscope, lavie, videocrafter1, videocrafter2, DeT}. These observations raise a fundamental question for the community: Does joint text-to-audio-video (T2AV) training improve video generation even when only the video output is evaluated? 
Answering this question matters. If the answer is negative, T2AV may primarily serve as a coordination mechanism that integrates existing T2V and text-to-audio (T2A) capabilities, yielding better synchrony without advancing the video model’s ceiling as a world model. Conversely, suppose hearing helps seeing, i.e., adding an audio denoising objective directly improves video learning. In that case, we may get a promising insight: Audio can act as a privileged signal that strengthens a model’s grasp of physical causality (e.g., contact $\rightarrow$ sound), thereby regularizing motion and interactions, as shown in~\cref{fig:teaser}. Intuitively, natural creatures rarely rely on vision alone: seeing a closed door does not reveal whether it will open next, but the sound of footsteps and a turning handle makes that prediction likely. From this perspective, audio is a key component for agents to understand the physical world.
While recent related open-sourced efforts (e.g., JavisDiT~\cite{JavisDiT}, UniVerse-1~\cite{UniVerse-1}) build capable T2AV systems with large A/V corpora, they primarily benchmark synchrony and overall generation quality. However, what remains missing is a controlled, head-to-head assessment that isolates the effect of joint A/V denoising and solely video denoising. In contrast, we address this gap through a systematic comparison of T2AV versus T2V under matched pre-trained models, data, and optimization settings, providing valuable insights for our conclusions.

To enable a clean study, we introduce \textbf{AVFullDiT}, a parameter-efficient framework that inherits knowledge from pre-trained T2V and T2A backbones. Its \textbf{AVFull-Attention} performs symmetric self-attention over concatenated audio and video tokens, encouraging rich cross-modal information exchange while preserving unimodal inductive biases. We further propose \textbf{AVSyncRoPE}, which rescales audio rotary positional encodings (RoPE) to align real-time token spacing with video, thereby improving video learning and, incidentally, A/V synchrony. 
Using two curated datasets—one smaller, category-focused set for fast iteration and a larger open-domain set for generalization—we train matched T2V and T2AV models and evaluate them on video-only evaluation sets. Our study yields three findings: \textbf{1) despite a slightly lower validation loss for T2V, T2AV consistently improves video metrics; 2) T2AV produces more realistic, tempered motion, avoiding exaggerated or frozen dynamics; and 3) T2AV enhances physical commonsense, yielding more plausible object interactions and mitigating failures such as contact avoidance.} We argue that these improvements arise from the world knowledge embedded in the audio modality, and our experimental results in~\cref{sec:exp} offer supporting evidence for this.
Our contributions can be summarized as:
\begin{itemize}
    \item We pose and systematically answer a fundamental question: Does hearing help seeing for video generation?
    \item We present AVFullDiT, with AVFull-Attention and AVSyncRoPE, to efficiently co-train audio and video.
    \item Through extensive controlled experiments, we demonstrate that audio supervision enhances video generation, clarifying where and how T2AV improves T2V performance and providing actionable insights for future multimodal world models.
\end{itemize}
\section{Related Work}

\noindent
\textbf{Multimodal co-training mutual benefit.}
Recent work has extended LLMs~\cite{Qwen, Qwen2, Qwen2.5, Qwen3, Deepseek-v3} to Vision–Language Models (VLMs) for perception tasks such as image captioning~\cite{LLaVA, Minigpt-4, InternVL, Qwen2.5-VL, Tarsier2, Sa2VA}. In parallel, diffusion-based generative models have triggered rapid advances in visual synthesis and editing~\cite{ldm, imagen, videocrafter1, videocrafter2, animatediff, lavie, modelscope, zeroscope, HunyuanVideo, Wan, CogVideoX}.
A growing body of research is investigating unified architectures that co-train understanding and generation, expecting potential mutual benefits~\cite{Bagel, Show-o2, Qwen-Image, Janus-Pro, Emu3.5, muddit}. Empirically, perception-oriented objectives enhance controllability and visual quality in visual generation and editing~\cite{Bagel, Show-o, Show-o2, Qwen-Image, Janus-Pro, Emu, Emu2, Emu3.5, Ming-Omni, Ming-Flash-Omni, CGG, LGVI, DiffSensei, DreamRelation}.
Beyond image–text, multimodal co-training has also been leveraged for video generation. VideoJAM~\cite{VideoJAM} demonstrates that pairing video synthesis with optical-flow joint denoising benefits scenes with significant motion, and UDPDiff~\cite{UDPDiff} shows that mask–video co-generation enhances generation fidelity. These findings suggest that multimodal co-training can induce positive transfer across modalities.
We argue that audio is a comparably important modality for enabling video generation systems to understand the world. However, prior work remains limited in this direction.
In this work, we present the first systematic investigation of audio–video co-training for T2V models, finding that cross-modal co-training delivers consistent gains and enhances the short-term capabilities of current video generation models.

\noindent
\textbf{Audio-video joint denoising.}
Recent closed-source systems Veo 3~\cite{Veo3} and Sora 2~\cite{Sora2} demonstrate striking advances in text-to-video quality, while jointly generating temporally aligned audio. These successes surface two questions for the community: (i) how to train audio–video generative models at comparable quality, and (ii) whether their superior video realism is attributable, in part, to the co-generation architecture itself.
Early explorations tackle the first question from different angles. SVG~\cite{SVG} and Wang et al.~\cite{Animate-Sound-Image} propose co-generation frameworks that leverage pre-trained T2V and T2A components and fine-tune on small datasets, demonstrating the feasibility of the task. In contrast, JavisDiT~\cite{JavisDiT} and UniVerse-1~\cite{UniVerse-1} develop large-scale data pipelines and training recipes to create general-purpose A/V generators.
While these efforts deliver compelling end-to-end results, they do not systematically address the second question: beyond synchrony, does introducing the audio modality directly enhance video generation quality?
Our work, to our knowledge, is the first to study this gap by isolating the effect of joint denoising versus a T2V-only objective. We present evidence that joint denoising yields consistent gains in video generation, indicating that the audio modality provides valuable information for video generation models.

\section{Method}

\begin{figure*}[t!]
	\centering
	\includegraphics[width=1.0\linewidth]{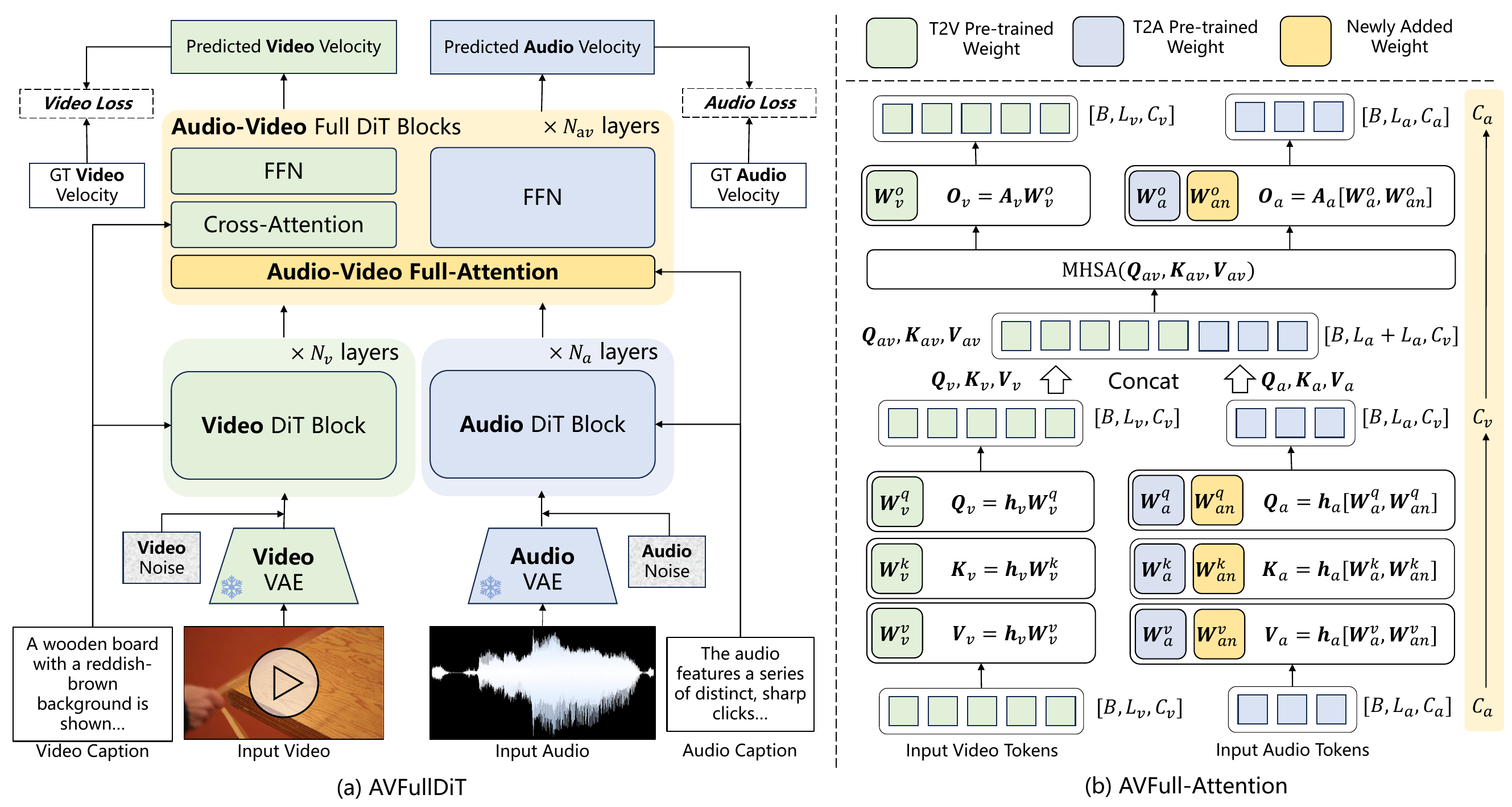}
	\caption{\textbf{The architecture of AVFullDiT and Audio-Video Full Attention}. (a) AVFullDiT reuses pre-trained T2V/T2A early towers and stacks joint blocks that predict video/audio velocities under a unified flow-matching loss. (b) AVFull-Attention performs symmetric MHSA over the concatenated audio–video token sequence using the video width as the joint dimension; audio projections are expanded with small adapter matrices. The attended sequence is split and projected back per modality.}
	\label{fig:method-architecture}
\end{figure*}

\begin{figure}[t!]
	\centering
	\includegraphics[width=1.0\linewidth]{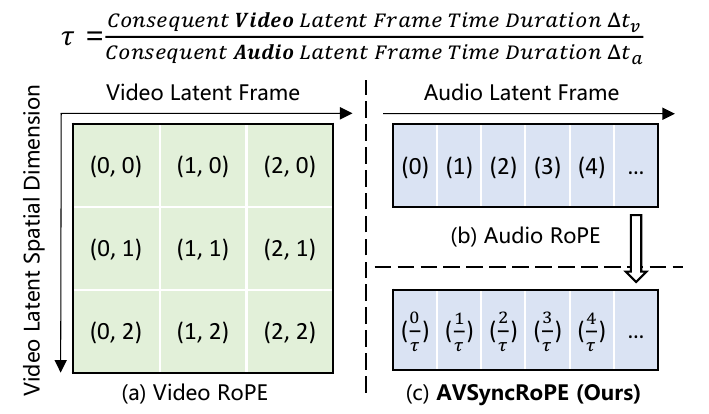}
	\caption{\textbf{The architecture of AVSyncRoPE}. (a–b) Vanilla RoPEs of the pre-trained video and audio DiTs live on different temporal scales. (c) We rescale audio positions so that video and audio tokens are aligned in real-time. This improves video learning, along with the side benefit of tighter A/V synchrony.}
	\label{fig:method-avsyncrope}
    \vspace{-5mm}
\end{figure}

In this section, we present AVFullDiT, an architecture that maximally reuses the weights of pre-trained T2V and T2A models while introducing minimal additional parameters. A new AVFull-Attention module bridges the two modalities, enabling rapid adaptation to a unified audio–video joint denoising objective. The overall architecture is shown in~\cref{fig:method-architecture}(a), and the AVFull-Attention design is detailed~\cref{fig:method-architecture}(b). To further enhance video quality, we introduce AVSyncRoPE, a simple but effective modification to vanilla audio RoPE that synchronizes audio and video along the token timeline. This synchronization enhances video learning, and as a secondary benefit, improves audio–video alignment. Its architecture is illustrated in~\cref{fig:method-avsyncrope}.

\subsection{AVFullDiT}

\noindent
\textbf{Overall architecture.}
As shown in \cref{fig:method-architecture}(a), AVFullDiT consists of two unimodal early towers, which are precisely the early layers from the pre-trained T2V and T2A models. In later layers, there are stacked audio–video joint blocks, which compute AVFull-Attention (detailed in~\cref{sec:avfull-attention}). We train the whole Transformer with a unified flow-matching velocity-prediction objective.

\noindent
\textbf{Latent representation and noising.}
Given an input video $\mathbf{D}_{v}$ and its paired audio $\mathbf{D}_{a}$, we first obtain latents with their corresponding pre-trained VAEs:
\begin{equation}
    \mathbf{x}_{v}^{0}=E_v(\mathbf{D}_{v}),\qquad
    \mathbf{x}_{a}^{0}=E_a(\mathbf{D}_{a}),
\end{equation}
where $E_v$ and $E_a$ are the video and audio VAE encoders, respectively. Then, for a sampled timestep $t$ and two Gaussian noises $\boldsymbol\epsilon_{v},\boldsymbol\epsilon_{a}\!\sim\!\mathcal{N}(\mathbf{0},\mathbf{I})$, we construct noisy latents by
\begin{equation}
    \mathbf{x}_{v}^{t}=(1 - t)\mathbf{x}_{v}^{0} + t\boldsymbol\epsilon_{v},\qquad
    \mathbf{x}_{a}^{t}=(1 - t)\mathbf{x}_{a}^{0} + t\boldsymbol\epsilon_{a},
\end{equation}
and define the velocity targets as
\begin{equation}
    \mathbf{v}_{v}^{t}=\boldsymbol\epsilon^{v} - \mathbf{x}_{v}^{0},\qquad
    \mathbf{v}_{a}^{t}=\boldsymbol\epsilon^{a} - \mathbf{x}_{a}^{0}.
\end{equation}
Additionally, we obtain video and audio caption representations $\mathbf{c}_{v}$ and $\mathbf{c}_{a}$ using frozen text encoders from the pre-trained video and audio checkpoints.

\noindent
\textbf{Unimodal DiT blocks on earlier layers.}
The noisy latents are synchronized with AVSyncRoPE along the time dimension (detailed in~\cref{sec:avsyncrope}). Tokens are then processed by stacks of unimodal DiT blocks, which are initialized from the pre-trained models:
\begin{equation}
    \begin{aligned}
        \mathbf{h}_{v}^{n} &= \mathcal{V}^{n}\!\left(\mathbf{h}_{v}^{n-1},\,\mathbf{c}_{v},\,t\right),\quad n \in \{1, 2, \dots, N_v\} \\
        \mathbf{h}_{a}^{n} &= \mathcal{A}^{n}\!\left(\mathbf{h}_{a}^{n-1},\,\mathbf{c}_{a},\,t\right),\quad n \in \{1, 2, \dots, N_a\}
    \end{aligned}
\end{equation}
where $n$ denotes the layer number. $\mathcal{V}_{n}$ and $\mathcal{A}_{n}$ are the pre-trained video and audio DiT blocks. $\mathbf{h}_{v}^{n}$ and $\mathbf{h}_{a}^{n}$ are the hidden states. $N_v$ and $N_a$ are the number of unimodal layers for each modality. We define them separately because the total number of layers for pre-trained video and audio DiTs is usually different. We set an equal number of later layers to compute the AVFull-Attention, leaving the earlier layers as unimodal. The unimodal earlier layers utilize pre-trained knowledge on each modality to obtain high-quality video and audio priors for further cross-modal fusion.

\noindent
\textbf{AVFull-Attention on later layers.}
\label{sec:avfull-attention}
On top of the two early towers, we stack $N_{av}$ Audio–Video Full DiT Blocks that exchange information across modalities via AVFull‑Attention. Specifically, we replace the Self-Attention for both pre-trained models with AVFull-Attention, leaving the Cross-Attention and FFN unchanged as unimodal.

Let $\mathbf{h}_v \in \mathbb{R}^{B\times L_v\times C_v}$ and $\mathbf{h}_a \in \mathbb{R}^{B\times L_a\times C_a}$ be the hidden states entering the AVFull-Attention, where $B$, $L_{(\cdot)}$, and $C_{(\cdot)}$ are batch size, sequence length, and channel dimension, respectively. We take the video channel width $C_v$ as the joint attention width.\footnote{In practice $C_v \ge C_a$ for common T2V/T2A models. For training stability, we choose to expand the audio channel.}
For the video branch, we reuse the pre‑trained T2V projections to get query, key, and value:
\begin{equation}
    \begin{aligned}
    \mathbf{Q}_v = \mathbf{h}_v\mathbf{W}^{q}_v,\quad
    \mathbf{K}_v &= \mathbf{h}_v\mathbf{W}^{k}_v,\quad
    \mathbf{V}_v = \mathbf{h}_v\mathbf{W}^{v}_v, \\
    \mathbf{Q}_v,\mathbf{K}_v,\mathbf{V}_v &\in \mathbb{R}^{B\times L_v\times C_v}.
    \end{aligned}
\end{equation}

Because the audio channel width $C_a$ may differ from $C_v$, we expand the audio projections to the same width by concatenating newly introduced, learnable projection matrices alongside the pre‑trained T2A weights:
\begin{equation}
    \begin{aligned}
        \mathbf{Q}_a&=\mathbf{h}_a\big[\mathbf{W}^{q}_a; \mathbf{W}^{q}_{an}\big],\\
        \mathbf{K}_a&=\mathbf{h}_a\big[\mathbf{W}^{k}_a; \mathbf{W}^{k}_{an}\big],\\
        \mathbf{V}_a&=\mathbf{h}_a\big[\mathbf{W}^{v}_a; \mathbf{W}^{v}_{an}\big],
    \end{aligned}
\quad
\mathbf{Q}_a,\mathbf{K}_a,\mathbf{V}_a\in\mathbb{R}^{B\times L_a\times C_v},
\end{equation}
where $\mathbf{W}^{(\cdot)}_a \in \mathbb{R}^{C_a\times C_a}$ are initialized from the T2A checkpoint and $\mathbf{W}^{(\cdot)}_{an} \in \mathbb{R}^{C_a\times (C_v-C_a)}$ are newly added parameters. $\big[\cdot\big]$ denotes concatenation along the channel dimension. This design enables the reuse of all pre-trained parameters while introducing only a minimal set of additional parameters to align channel sizes.

We now form a single sequence by concatenating audio and video tokens along the sequence dimension and perform standard multi‑head attention:
\begin{equation}
    \begin{aligned}
    \mathbf{Q}_{av}&=\mathrm{Concat}(\mathbf{Q}_v,\mathbf{Q}_a), \\
    \mathbf{K}_{av}&=\mathrm{Concat}(\mathbf{K}_v,\mathbf{K}_a), \\
    \mathbf{V}_{av}&=\mathrm{Concat}(\mathbf{V}_v,\mathbf{V}_a), \\
    \mathbf{A}_{av}&=\mathrm{MHSA}(\mathbf{Q}_{av},\mathbf{K}_{av},\mathbf{V}_{av}). \\
    \end{aligned}
\end{equation}
Because the attention is computed over the union of audio and video tokens, information flows bidirectionally and can propagate across the two modalities within a single operation. In the overall architecture, AVFull-Attention provides the core and only multimodal computation.

Finally, we split the attended sequence back to each modality and apply modality‑specific output projections:
\begin{equation}
    \mathbf{A}_v=\mathbf{A}_{av}[:,:L_v,:],\quad
    \mathbf{A}_a=\mathbf{A}_{av}[:,L_v:,:],
\end{equation}
\begin{equation}
    \mathbf{o}_v=\mathbf{A}_v\mathbf{W}^{o}_v,\qquad
    \mathbf{o}_a=\mathbf{A}_a,[\mathbf{W}^{o}_a;\mathbf{W}^{o}_{an}],
\end{equation}
where $\mathbf{W}^{o}_v \in \mathbb{R}^{C_v\times C_v}$ is initialized from the T2V checkpoint, $\mathbf{W}^{o}_a \in \mathbb{R}^{C_a\times C_a}$ comes from T2A, and $\mathbf{W}^{o}_{an} \in \mathbb{R}^{(C_v-C_a)\times C_a}$ is newly added. Hence, we transfer the audio latent back into its original channel width $C_a$.

After $N_{av}$ AVFull blocks, the model predicts video and audio velocities $\tilde{\mathbf{v}}_v$ and $\tilde{\mathbf{v}}_a$. The loss is computed as:
\begin{equation}
    \begin{aligned}
        \mathcal{L}_v=(\tilde{\mathbf{v}}_v - \mathbf{v}_v)^2,\quad&
        \mathcal{L}_a=(\tilde{\mathbf{v}}_a - \mathbf{v}_a)^2, \\
        \mathcal{L} = \lambda_v\mathcal{L}_v &+ \lambda_a\mathcal{L}_a,
    \end{aligned}
\end{equation}
where $\lambda_v$ and $\lambda_a$ are video and audio loss weights.

\noindent
\textbf{Discussion.}
Compared with modeling cross-modal information by Cross‑Attentions, AVFull‑Attention has three advantages: (i) symmetry: audio and video participate in the same attention graph; (ii) multi‑hop fusion: tokens from one modality can interact via tokens of the other within the same layer; and (iii) parameter efficiency: only a small set of adapter weights ${\mathbf{W}^{q}_{an},\mathbf{W}^{k}_{an},\mathbf{W}^{v}_{an},\mathbf{W}^{o}_{an}}$ is added to align channel sizes, while all pre‑trained weights are reused and fine‑tuned end‑to‑end.

\subsection{AVSyncRoPE}
\label{sec:avsyncrope}

\noindent
\textbf{Motivation.}
The pre-trained T2V and T2A DiTs utilize RoPEs that are defined in their own latent spaces. Because those latent frames cover different time durations, the two positional systems are on different time scales. When we concatenate audio and video tokens for AVFull‑Attention, a token pair that occurs at the same real time can carry very different RoPE phases, which makes it ambiguous for the model to learn which audio token should align with which video token in the real time dimension.

\noindent
\textbf{AVSyncRoPE.}
Let $\Delta t_v$ and $\Delta t_a$ denote the duration covered by the consequent video and audio latent frame, respectively (determined by the VAEs’ temporal scale factors, video fps, and audio sampling rate). We define the constant
\begin{equation}
\tau = \frac{\Delta t_v}{\Delta t_a}.
\end{equation}
Let $\mathbf{r} = \mathrm{RoPE}(p)$ denote the standard RoPE that rotates the time dimension by angles proportional to the temporal position index $p$. In the pre-trained models, integer indices $p \in \{0, 1, \dots\}$ are used: $p_v$ for video and $p_a$ for audio.
AVSyncRoPE is a parameter‑free, drop‑in change that rescales the audio position index so that both modalities are expressed in the same time scales. Concretely, we divide the audio index by $\tau$:
\begin{equation}
    \begin{aligned}
        \mathbf{r}_v &= \mathrm{RoPE}(p_v), \quad
        \mathbf{r}_a = \mathrm{RoPE}(\frac{p_a}{\tau}).
    \end{aligned}
\label{eq:avsyncrope}
\end{equation}
After rescaling, a video token at index $p$ and an audio token at index $\tau p$ share the same RoPE value, which corresponds to the fact that they are at the same point in time. We apply AVSyncRoPE among all attention layers in the unimodal audio tower and in the AVFull‑Attention blocks.

\subsection{Data Collection}
\label{sec:method-data-collection}

\begin{table}[t]
    \centering
    \caption{\textbf{Datasets used in experiments.} \#Clips and Duration are those after pre-processing.}
    \label{tab:method-dataset}
    \scalebox{0.55}{
    \begin{tabular}{lcccc}
\toprule
Dataset & Source & \#Clips & Duration (s) & Domain \\
\midrule
AVSync15~\cite{AVSync15} & VGGSound~\cite{VGGSound} & 1,499 & 2 & Motion and Related Sound Effects \\
Landscape~\cite{Landscape} & Youtube & 778 & 2 & Nature Scenes and Sounds \\
The Greatest Hits~\cite{TheGreatestHits} & Self-Collected & 977 & 2 & Clicking Various Surfaces \\
VGGSound~\cite{VGGSound} & Youtube & 107,648 & 5 & Open-World Sounded Video \\
\bottomrule
    \end{tabular}}
\end{table}

\noindent
\textbf{Source.}
To thoroughly evaluate the influence of the T2AV training diagram on T2V models, we curate a heterogeneous corpus that balances open‑domain everyday scenes, natural environments, and human performance content. Specifically, we draw from four sources: AVSync15~\cite{AVSync15}, a curated set that emphasizes tightly synchronized audiovisual events. Landscape~\cite{Landscape}: nature/ambient footage with slowly varying dynamics. The Greatest Hits~\cite{TheGreatestHits}: videos containing a stick clicking on various surfaces, which is very compact in motion and sound. And VGGSound~\cite{VGGSound}: open‑domain audio–video clips. This is a large dataset for training at scale. This mixture exposes the model to a wide range of motion statistics, acoustic textures, and event densities, which we think is adequate for stable and convincing conclusions. The statistics are shown in~\cref{tab:method-dataset}.

\noindent
\textbf{Pre-precessing and annotation.}
After downloading the full raw dataset, we apply data pre-processing to remove degenerate content. Specifically, we apply (i) duplicate video removal using metadata; (ii) silence video removal based on average audio volume; and (iii) portrait resolution video removal.
To maximize reuse of pre‑trained unimodal token–text interactions in our T2V and T2A models, we annotate two captions per clip: a silent video‑focused description $\mathbf{c}_v$ and an audio‑focused description $\mathbf{c}_a$. Video captions are generated using Tarsier2~\cite{Tarsier2}, which is prompted with silent videos only, aiming to describe visible entities, actions, camera motion, and scene context. 
Audio captions are produced with Qwen2.5‑Omni~\cite{Qwen2.5-Omni}, prompted with video and audio. It describes the audio while ignoring purely visual attributes. The average token count for video and audio captions is approximately 116 and 47, which is suitable for the pre-trained T2V and T2A text encoder checkpoints.

\section{Experiments}
\label{sec:exp}

\subsection{Settings}

\noindent
\textbf{Dataset.}
Based on the curated corpus described in~\cref{sec:method-data-collection}, we partition the data into two distinct datasets. The first, a smaller and more specialized dataset we term ALT-Merge, aggregates clips from AVSync15~\cite{AVSync15}, Landscape~\cite{Landscape}, and The Greatest Hits~\cite{TheGreatestHits}. We set aside a total of 300 clips for evaluation, ensuring a balanced contribution of approximately 100 clips from each of the three sources. The remaining 2,954 clips constitute the training set. The second utilizes the large-scale VGGSound dataset~\cite{VGGSound}. We split 106,717 clips for training and 731 clips for evaluation. For the evaluation set, we perform balanced sampling across their semantic categories.

\noindent
\textbf{Implementation details.}
We utilize Wan2.2-TI2V-5B~\cite{Wan} as the pre-trained video diffusion model and TangoFlux~\cite{TangoFlux} as the pre-trained audio diffusion model. These models were selected for their academically manageable scale, state-of-the-art performance, and their shared use of a flow matching objective. During training, we fine-tune all parameters of AVFullDiT while keeping the VAEs and text encoders frozen. A consistent set of hyperparameters is used across all experiments, featuring a learning rate of 1e-5 and the AdamW optimizer~\cite{AdamW}. All models are trained at a fixed resolution of 192 $\times$ 320. For the ALT-Merge dataset, we train for 7,500 steps with a batch size of 8, using video clips of 2-second duration. For the VGGSound dataset, training is extended to 100,000 steps with a batch size of 24, using 5-second video clips to ensure joint convergence of both audio and video. Training was conducted on NVIDIA H20 141G GPUs. In inference, we use 50 denoising steps. For classifier-free guidance (CFG) in the T2AV model, the positive branch is conditioned on both video and audio text prompts, while the negative branch is conditioned on negative prompts for both modalities. The guidance scales for video and audio are set to 5.0 and 4.5, respectively, inheriting the settings from the pre-trained models.

\noindent
\textbf{Metrics.}
To comprehensively evaluate the capabilities of video generation, we employ a suite of automated metrics. For general video quality, we adopt five dimensions from VBench~\cite{VBench}: Background Consistency, which assesses the stability of static regions; Dynamic Degree, measuring the magnitude of motion; Image Quality, for the aesthetic appeal of frames; Subject Consistency, ensuring the coherent appearance of primary subjects over time; and Text Consistency, which evaluates the semantic alignment with the silent video prompt.
To test our central hypothesis that the audio modality aids in understanding the physical world, we incorporate a metric for physical commonsense. Specifically, we use the pre-trained model from Videophy-2~\cite{Videophy-2} to score the physical plausibility of generated content.
In our ablation studies, we introduce additional metrics for audio quality and audio-video synchrony. Audio fidelity is measured by Fréchet Audio Distance (FAD)~\cite{FAD}. We use the CLAP~\cite{CLAP} score to evaluate the semantic similarity between the generated audio and the audio prompt. Finally, a pre-trained Synchformer model~\cite {Synchformer} is utilized to calculate the temporal offset between the audio and video tracks.

\subsection{Quantitative Results}

\begin{table*}[!t]
\centering
\caption{\textbf{Video generation comparison between T2AV and T2V training}. Improvements are highlighted in green, while decreases are highlighted in red. Metrics with a $^\dagger$ are not definitely better when higher. Numbers with the largest percentage improvements are bolded.}
\label{tab:exp-t2av-vs-t2v}
\scalebox{0.70}{
\begin{tabular}{l|l|l|cccccc}
\toprule
\textbf{Method} & \textbf{Dataset} & \textbf{Evaluation Set} & \textbf{BGConsis} & \textbf{Dynamic}$^\dagger$ & \textbf{ImageQual} & \textbf{SubjQual} & \textbf{TextConsis} & \textbf{Physics} \\
\midrule
\multirow{8}{*}{T2V} & \multirow{4}{*}{ALT-Merge} & Full Set & 97.44 & 52.67 & 59.04 & 95.79 & 9.91 & 4.27 \\
& & \textbar \textemdash \ AVSync15~\cite{AVSync15} & 98.14 & 45.00 & 61.10 & 97.69 & 9.78 & 4.25 \\
& & \textbar \textemdash \ Landscape~\cite{Landscape} & 98.11 & 16.25 & 54.29 & 96.59 & 8.68 & 4.85 \\
& & \raisebox{0.4ex}{$\llcorner$} \textemdash \ TheGreatestHits~\cite{TheGreatestHits} & 96.05 & 91.00 & 60.39 & 92.87 & 11.05 & 3.82 \\
\cmidrule{2-9}
& \multirow{3}{*}{VGGSound~\cite{VGGSound}} & Full Set & 94.19 & 77.70 & 59.57 & 88.72 & 9.34 & 4.34 \\
& & \textbar \textemdash \ AV-Tight & 94.44 & 79.71 & 60.18 & 89.07 & 9.62 & 3.99 \\
& & \raisebox{0.4ex}{$\llcorner$} \textemdash \ AV-Loose & 94.06 & 77.08 & 59.68 & 88.55 & 9.24 & 4.43 \\
\midrule
\multirow{8}{*}{T2AV} & \multirow{4}{*}{ALT-Merge} & Full Set & 97.93 (\textcolor{Green}{+0.50\%}) & 49.67 (\textcolor{Red}{-5.70\%}) & 59.87 (\textcolor{Green}{+1.40\%}) & 96.30 (\textcolor{Green}{+0.53\%}) & 10.00 (\textcolor{Green}{+0.91\%}) & 4.36 (\textcolor{Green}{+2.11\%}) \\
& & \textbar \textemdash \ AVSync15 & 98.42 (\textcolor{Green}{+0.28\%}) & 40.00 (\textcolor{Red}{-11.11\%}) & \textbf{62.85 (\textcolor{Green}{+2.86\%})} & 97.88 (\textcolor{Green}{+0.19\%}) & 9.98 (\textcolor{Green}{+2.04\%}) & 4.36 (\textcolor{Green}{+2.59\%}) \\
& & \textbar \textemdash \ Landscape & 98.33 (\textcolor{Green}{+0.22\%}) & 31.25 (\textcolor{Green}{+92.31\%}) & 54.26 (\textcolor{Red}{-0.01\%}) & 97.10 (\textcolor{Green}{+0.53\%}) & \textbf{8.95 (\textcolor{Green}{+3.11\%})} & 4.90 (\textcolor{Green}{+1.03\%}) \\
& & \raisebox{0.4ex}{$\llcorner$} \textemdash \ TheGreatestHits & \textbf{97.01 (\textcolor{Green}{+1.00\%})} & 76.00 (\textcolor{Red}{-16.48\%}) & 60.78 (\textcolor{Green}{+0.64\%}) & \textbf{93.78 (\textcolor{Green}{+0.98\%})} & 10.86 (\textcolor{Red}{-1.72\%}) & \textbf{3.94 (\textcolor{Green}{+3.14\%})} \\
\cmidrule{2-9}
& \multirow{3}{*}{VGGSound} & Full Set & 94.27 (\textcolor{Green}{+0.08\%}) & 77.98 (\textcolor{Green}{+0.36\%}) & 58.78 (\textcolor{Red}{-1.33\%}) & 88.92 (\textcolor{Green}{+0.22\%}) & 9.58 (\textcolor{Green}{+2.57\%}) & 4.37 (\textcolor{Green}{+0.69\%}) \\
& & \textbar \textemdash \ AV-Tight & \textbf{94.67 (\textcolor{Green}{+0.24\%})} & 78.98 (\textcolor{Red}{-0.92\%}) & \textbf{60.93 (\textcolor{Green}{+1.25\%})} & \textbf{89.68 (\textcolor{Green}{+0.68\%})} & \textbf{9.88 (\textcolor{Green}{+2.70\%})} & \textbf{4.09 (\textcolor{Green}{+2.51\%})} \\
& & \raisebox{0.4ex}{$\llcorner$} \textemdash \ AV-Loose & 94.17 (\textcolor{Green}{+0.12\%}) & 77.44 (\textcolor{Green}{+0.47\%}) & 58.45 (\textcolor{Red}{-2.06\%}) & 88.77 (\textcolor{Green}{+0.25\%}) & 9.46 (\textcolor{Green}{+2.38\%}) & 4.44 (\textcolor{Green}{+0.22\%}) \\
\bottomrule
 \end{tabular}}
\end{table*}

\begin{figure}[t!]
	\centering
	\includegraphics[width=1.0\linewidth]{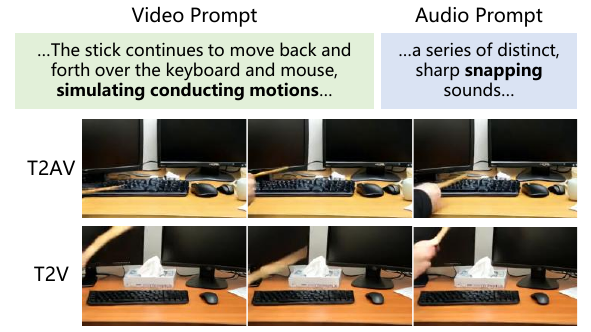}
	\caption{Example of wrong video prompt annotation in TheGreatestHits. The bolded video prompts are the incorrect part, while the audio prompts indicate the correct generation result for T2AV.}
	\label{fig:exp-TheGreatestHits-annotation}
\end{figure}

\noindent
\textbf{Metrics comparison.}
We conduct a quantitative comparison between the T2AV model and the T2V counterpart under identical training settings. To further validate our findings, we also evaluate performance on representative subsets curated from each evaluation set. The comprehensive results are presented in~\cref{tab:exp-t2av-vs-t2v}.

On the full ALT-Merge evaluation set, the T2AV model demonstrates consistent improvements over the T2V baseline across all metrics except for Dynamic Degree. This provides strong initial evidence that joint audio-video training benefits the video generation process. It is important to note that a higher Dynamic Degree, which measures the magnitude of motion, does not necessarily correlate with better video quality. We observe that the T2V model exhibits higher variance in this metric, often producing videos with either exaggerated motion or excessive stillness. In contrast, the T2AV model generates motion that is more stable and realistic, with a more concentrated distribution for the Dynamic Degree score. This is evident in the subset results: on The Greatest Hits, a dataset featuring rapid stick-hitting motions, T2V scores 91.00 while T2AV scores a more tempered 76.00. Conversely, on the less dynamic Landscape dataset, T2V scores a low 16.25, whereas T2AV achieves 31.25, indicating a tendency towards more plausible motion levels. These nuances will be further detailed in~\cref{sec:qualitative}.
The most significant gains are observed in the Physics metric, which measures physical commonsense. On the motion- and contact-intensive subset of The Greatest Hits, the T2AV model achieves the largest improvement of 3.14\%. The AVSync15 and Landscape subsets show gains of 2.59\% and 1.03\%, respectively. This trend strongly suggests that for scenes involving substantial motion and object interactions, joint training with audio helps the model better grasp real-world physical laws, validating our core hypothesis that ``hearing helps seeing.''
Another noteworthy result is the lower Text Consistency on The Greatest Hits. Upon inspection, we find this is attributable to systematic errors in the video prompts generated by the Tarsier2~\cite{Tarsier2} model for silent videos. As shown in~\cref{fig:exp-TheGreatestHits-annotation}, the silent video captioner often mislabels a stick striking an object as a stick merely waving in the air, as it cannot perceive the acoustic cues of impact. Our T2AV model, however, correctly leverages the audio prompt describing ``snapping sounds'' to generate the appropriate contact-rich video. This explains the lower score and highlights a key weakness of the T2V training paradigm: its susceptibility to inaccuracies in silent video annotation. Notably, even with accurate prompts, the T2V model often generates videos that avoid contact, a failure mode we demonstrate in~\cref{sec:qualitative}.

On the large-scale VGGSound dataset, our findings are consistent. The T2AV model outperforms the T2V baseline on most metrics in the full set, confirming that the introduction of an audio objective enhances performance in the general domain. To investigate whether this improvement stems from learned real-world correlations, we partitioned the dataset into two subsets: AV-Tight (e.g., chopping wood, church bell ringing), where audio and visual events are highly coupled, and AV-Loose, where they are not. The results in~\cref{tab:exp-t2av-vs-t2v} show that the AV-Tight subset not only achieves the highest scores but also registers the most significant percentage improvements across all metrics (excluding Dynamic), particularly in Text Consistency (+2.70\%) and Physics (+2.51\%). This strongly supports the hypothesis that the model's enhanced understanding is derived from learning the intrinsic coupling between modalities as they occur in the physical world.

\begin{figure}[ht!]
    \centering
    \begin{subfigure}{0.49\linewidth}
        \centering
        \includegraphics[width=\textwidth]{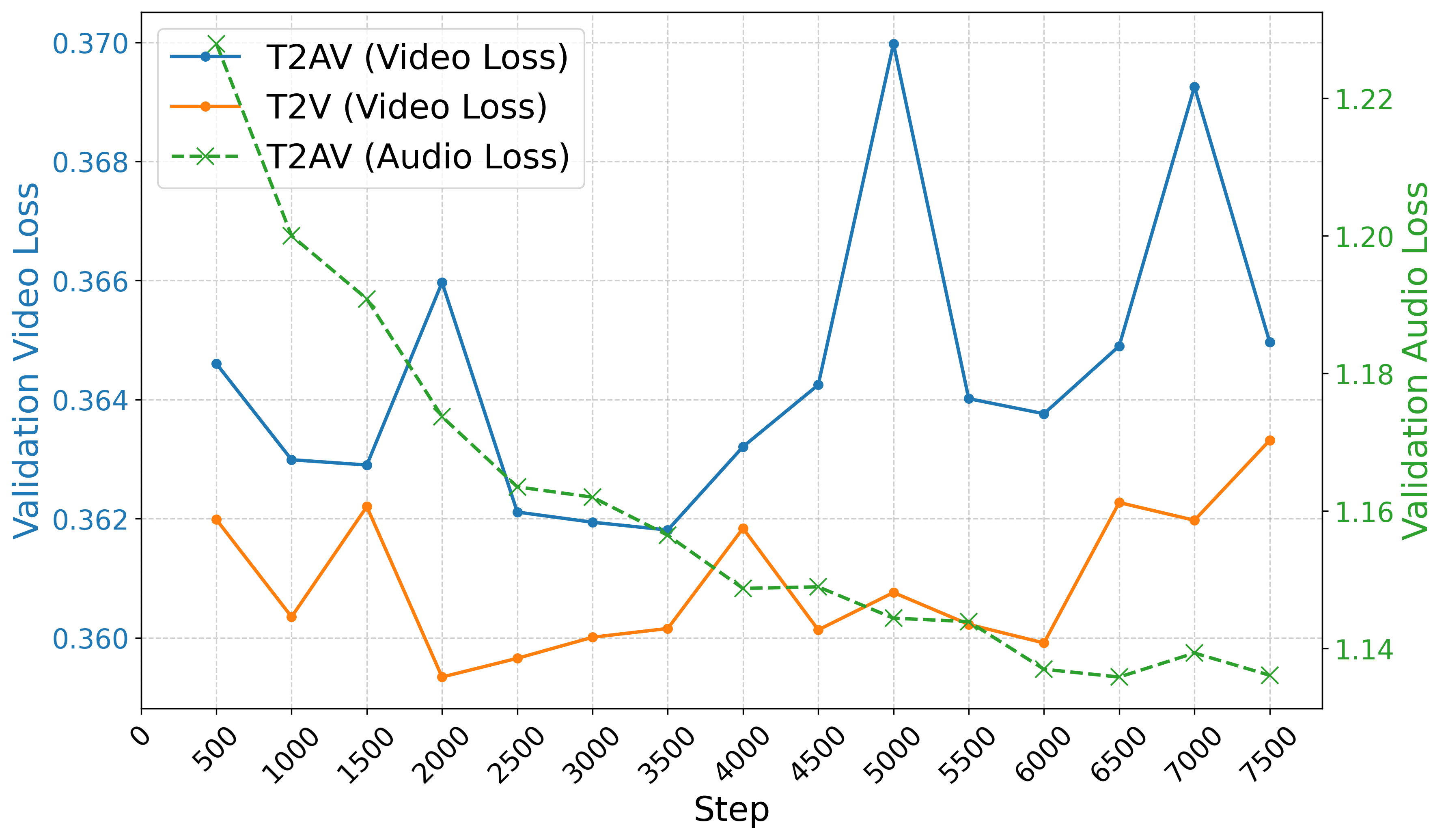}
        \caption{Validation loss on ALT-Merge.}
        \label{fig:exp-valloss-t2av-vs-t2v-ALT-Merge}
    \end{subfigure}
    \hfill
    \begin{subfigure}{0.49\linewidth}
        \centering
        \includegraphics[width=\textwidth]{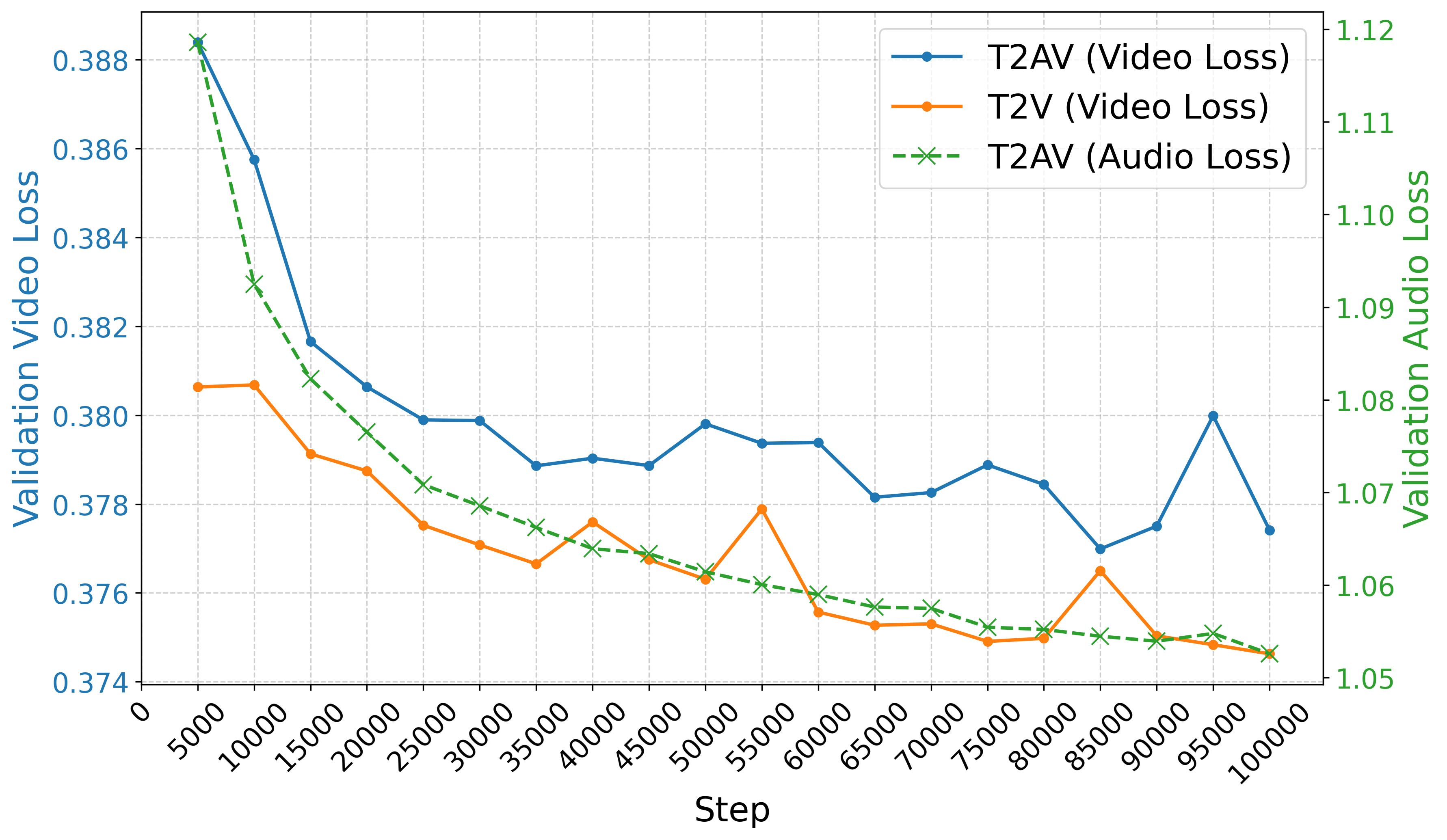}
        \caption{Validation loss on VGGSound.}
        \label{fig:exp-valloss-t2av-vs-t2v-VGGSound}
    \end{subfigure}
    \caption{\textbf{Validation loss comparison between T2AV and T2V.}}
    \label{fig:exp-valloss-t2av-vs-t2v}
\end{figure}

\noindent
\textbf{Validation loss comparison.}
Beyond the metrics, we also track the validation loss, as illustrated in~\cref{fig:exp-valloss-t2av-vs-t2v}. We observe a consistent trend across both the ALT-Merge and VGGSound datasets: the video validation loss for the T2V-only model remains slightly lower than that of the T2AV model, with the difference being less than 0.01. This finding suggests that a slightly lower validation loss does not directly equate to superior generative quality. Furthermore, we note that the audio loss in the T2AV model converges more slowly than the video loss. We attribute this to the newly introduced parameters in the audio branch, which must be learned from scratch. This outcome aligns with our design to prioritize video performance, leveraging the audio modality as a regularizer to enhance video generation.

\subsection{Qualitative Results}
\label{sec:qualitative}

\begin{figure*}[t!]
	\centering
	\includegraphics[width=1.0\linewidth]{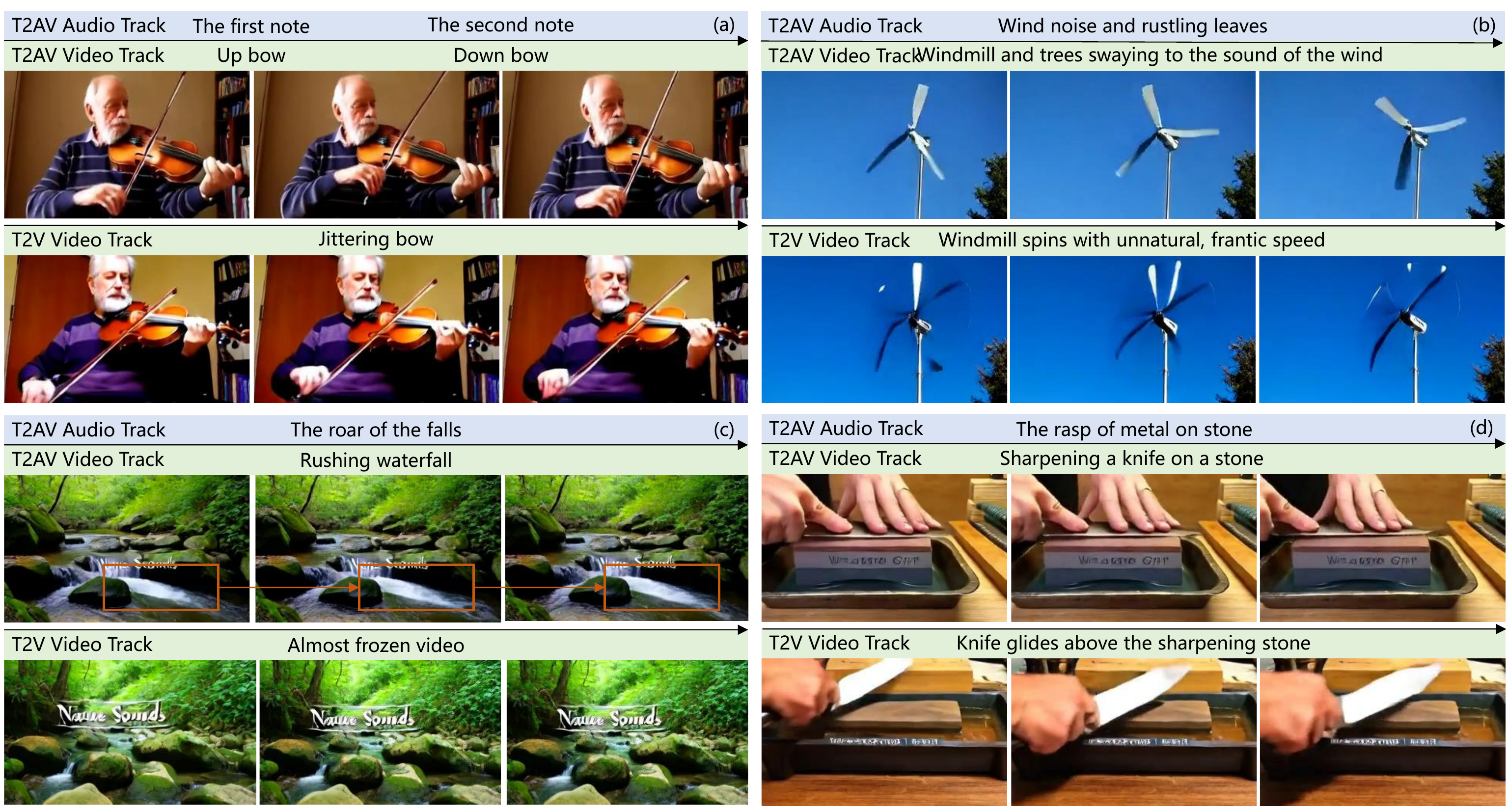}
	\caption{\textbf{Qualitative comparison between T2AV and T2V.} Video and audio track content are described using text. Motion in (c) is highlighted for clarity. Please refer to the appendix for real audio-video results.}
	\label{fig:exp-qualitative}
\end{figure*}

\cref{fig:exp-qualitative} presents the qualitative comparison. The results visually substantiate the high variance observed in the T2V model's Dynamic Degree metric. For instance, when prompted to generate a person playing the violin (a) or a windmill in the wind (b), the T2V model produces motion that is either unnaturally agitated (jittering bow) or physically implausible (frantically spinning blades). In contrast, the T2AV model generates more controlled and realistic actions, such as a distinct up-and-down bow stroke and a reasonably paced windmill rotation, which are coherently synchronized with corresponding audio cues. Similarly, in the waterfall scene (c), T2AV renders dynamic water flow, while the T2V generation is nearly static. These cases demonstrate that T2AV produces more stable and natural movements, thereby avoiding the extremes of exaggerated or minimal motion commonly found in the T2V baseline.
In the knife-sharpening example (d), the T2V model fails to depict contact, showing the knife gliding above the stone. This ``contact avoidance'' is a known failure mode for video generation models~\cite{animatediff, videocrafter1, videocrafter2, zeroscope, lavie}. The T2AV model, however, correctly generates the physical interaction, with the blade making contact with the stone, accompanied by a synchronized metallic scraping sound. This suggests that the audio signal acts as a strong regularizer, encouraging the model to generate videos that are more consistent with the physical laws of our world.
Given that the motion and audio cannot be fully conveyed through static images, \textbf{we strongly recommend viewing the complete audio-video results provided in the appendix.}

\subsection{Ablation Study}

\begin{table}[!ht]
    \centering
    \caption{\textbf{Ablation studies.} IQ, SC, and PH are Image Quality, Subject Consistency, and Physics. FAD is in units of $10^4$. All ablations are done on ALT-Merge.}
    \begin{subtable}{1.0\linewidth}
        \centering
        \caption{T2AV cross-modal attention.}
        \label{tab:exp-ablation-attn}
        \resizebox{1.0\linewidth}{!}{
        \begin{tabular}{l|cccccc}
            \toprule
            \textbf{Method} & FAD $\downarrow$ & CLAP $\uparrow$ & IQ $\uparrow$ & SC $\uparrow$ & PH $\uparrow$ & Sync $\downarrow$ \\
            \midrule
            Cross-Attention & 11.37 & 27.29 & 58.54 & 95.68 & 4.35 & \textbf{0.27} \\
            \textbf{AVFull-Attention (Ours)} & \textbf{9.36} & \textbf{38.75} & \textbf{59.87} & \textbf{96.30} & \textbf{4.36} & 0.29 \\
            \bottomrule
        \end{tabular}}
    \end{subtable}
    \begin{subtable}{1.0\linewidth}
        \centering
        \caption{T2AV RoPE design.}
        \label{tab:exp-ablation-avsyncrope}
        \resizebox{1.0\linewidth}{!}{
        \begin{tabular}{l|cccccc}
            \toprule
            \textbf{Method} & FAD $\downarrow$ & CLAP $\uparrow$ & IQ $\uparrow$ & SC $\uparrow$ & PH $\uparrow$ & Sync $\downarrow$ \\
            \midrule
            Vanilla & 8.93 & \textbf{44.46} & 57.76 & 94.05 & 4.22 & 0.32 \\
            Expand Video & \textbf{8.81} & 43.09 & 56.34 & 93.57 & 4.20 & \textbf{0.26} \\
            \textbf{Shrink Audio (Ours)} & 9.36 & 38.75 & \textbf{59.87} & \textbf{96.30} & \textbf{4.36} & 0.29 \\
            \bottomrule
        \end{tabular}}
    \end{subtable}
\end{table}

\noindent
\textbf{Cross-modal attention mechanism.} We ablate the modeling of audio-video interaction. As an alternative to our proposed AVFull-Attention, we designed a Cross-Attention baseline. In this setup, following the unimodal self-attention in the later $N_{av}$ layers, a cross-attention module is added, where each modality receives the other as its key-value input. As shown in~\cref{tab:exp-ablation-attn}, AVFull-Attention surpasses the Cross-Attention baseline on all audio and video quality metrics, with only a minor trade-off in the synchronization score. This result demonstrates that AVFull-Attention is a more effective approach for learning audio-video joint denoising. We hypothesize this is because AVFull-Attention maximally preserves the architecture of the pre-trained models, elegantly and symmetrically unifying them through a single, powerful attention operation.

\noindent
\textbf{AVSyncRoPE design.} We ablate the design of AVSyncRoPE. An alternative to shrinking the audio RoPE is to expand the video RoPE to match the audio's. \cref{tab:exp-ablation-avsyncrope} compares these two approaches against the vanilla RoPE baseline. The results on video-related metrics (Image Quality, Subject Consistency, and Physics) show that our proposed method is clearly superior. Conversely, the ``Expand Video'' method performs best on audio metrics, followed by the vanilla baseline. We infer that modifying a modality's RoPE shifts its features away from the distribution familiar to the pre-trained model, thereby degrading that modality's performance. To maximize video quality, our primary objective, we choose to shrink the audio RoPE. Notably, both the ``Expand Video'' and ``Shrink Audio'' variants achieve significantly better synchronization than the vanilla baseline, confirming the validity of our motivation: to temporally align the audio and video positional encodings.

\section{Conclusion}

In this work, we pose a fundamental question: Does hearing help seeing? Specifically, we investigated whether joint audio-video denoising offers benefits to video generation. To systematically address this, we introduced AVFullDiT, a parameter-efficient architecture designed to maximize the leverage of knowledge from pre-trained T2V and T2A models. Our design incorporates a novel AVFull-Attention mechanism and an AVSyncRoPE module, both validated for their effectiveness.
Through rigorous experiments on two datasets of varying size and distribution, we obtain three main conclusions. \textbf{1) Despite a marginally higher video validation loss, T2AV training consistently enhances video generation across multiple metrics. 2) The T2AV model generates motion that is more realistic and tempered, avoiding the extremes of exaggerated or static scenes. 3) The T2AV model demonstrates an improved understanding of physical commonsense, more reliably generating plausible object interactions and avoiding common failure modes, such as ``contact avoidance.''}
In summary, our findings confirm that hearing does help seeing. Audio-video cross-modal joint denoising yields consistent improvements for video generation, paving the way for the development of stronger multimodal world models.

\noindent
\textbf{Acknowledgement.} This work is supported by the National Key Research and Development Program of China (No. 2023YFC3807600).
{
    \small
    \bibliographystyle{ieeenat_fullname}
    \bibliography{main}

@String(CVPR= {IEEE Conf. Comput. Vis. Pattern Recog.})

@String(ICCV= {Int. Conf. Comput. Vis.})

@String(ECCV= {Eur. Conf. Comput. Vis.})

@String(ICASSP=	{ICASSP})

@String(ICLR = {Int. Conf. Learn. Represent.})

@String(CVPR  = {CVPR})

@String(ICCV  = {ICCV})

@String(ECCV  = {ECCV})

@String(ICLR  = {ICLR})

@article{Qwen,
  title={Qwen technical report},
  author={Bai, Jinze and Bai, Shuai and Chu, Yunfei and Cui, Zeyu and Dang, Kai and Deng, Xiaodong and Fan, Yang and Ge, Wenbin and Han, Yu and Huang, Fei and others},
  journal={arXiv preprint arXiv:2309.16609},
  year={2023}
}

@article{Qwen2,
  title={Qwen2 technical report},
  author={Team, Qwen and others},
  journal={arXiv preprint arXiv:2407.10671},
  year={2024}
}

@article{Qwen2.5,
  title={Qwen2. 5-coder technical report},
  author={Hui, Binyuan and Yang, Jian and Cui, Zeyu and Yang, Jiaxi and Liu, Dayiheng and Zhang, Lei and Liu, Tianyu and Zhang, Jiajun and Yu, Bowen and Lu, Keming and others},
  journal={arXiv preprint arXiv:2409.12186},
  year={2024}
}

@article{Qwen3,
  title={Qwen3 technical report},
  author={Yang, An and Li, Anfeng and Yang, Baosong and Zhang, Beichen and Hui, Binyuan and Zheng, Bo and Yu, Bowen and Gao, Chang and Huang, Chengen and Lv, Chenxu and others},
  journal={arXiv preprint arXiv:2505.09388},
  year={2025}
}

@article{Deepseek-v3,
  title={Deepseek-v3 technical report},
  author={Liu, Aixin and Feng, Bei and Xue, Bing and Wang, Bingxuan and Wu, Bochao and Lu, Chengda and Zhao, Chenggang and Deng, Chengqi and Zhang, Chenyu and Ruan, Chong and others},
  journal={arXiv preprint arXiv:2412.19437},
  year={2024}
}

@article{LLaVA,
  title={Visual instruction tuning},
  author={Liu, Haotian and Li, Chunyuan and Wu, Qingyang and Lee, Yong Jae},
  journal={NeurIPS},
  year={2023}
}

@article{Minigpt-4,
  title={Minigpt-4: Enhancing vision-language understanding with advanced large language models},
  author={Zhu, Deyao and Chen, Jun and Shen, Xiaoqian and Li, Xiang and Elhoseiny, Mohamed},
  journal={arXiv preprint arXiv:2304.10592},
  year={2023}
}

@inproceedings{InternVL,
  title={Internvl: Scaling up vision foundation models and aligning for generic visual-linguistic tasks},
  author={Chen, Zhe and Wu, Jiannan and Wang, Wenhai and Su, Weijie and Chen, Guo and Xing, Sen and Zhong, Muyan and Zhang, Qinglong and Zhu, Xizhou and Lu, Lewei and others},
  booktitle={CVPR},
  year={2024}
}

@article{Qwen2.5-VL,
  title={Qwen2. 5-vl technical report},
  author={Bai, Shuai and Chen, Keqin and Liu, Xuejing and Wang, Jialin and Ge, Wenbin and Song, Sibo and Dang, Kai and Wang, Peng and Wang, Shijie and Tang, Jun and others},
  journal={arXiv preprint arXiv:2502.13923},
  year={2025}
}

@article{Tarsier2,
  title={Tarsier2: Advancing large vision-language models from detailed video description to comprehensive video understanding},
  author={Yuan, Liping and Wang, Jiawei and Sun, Haomiao and Zhang, Yuchen and Lin, Yuan},
  journal={arXiv preprint arXiv:2501.07888},
  year={2025}
}

@article{Sa2VA,
  title={Sa2va: Marrying sam2 with llava for dense grounded understanding of images and videos},
  author={Yuan, Haobo and Li, Xiangtai and Zhang, Tao and Huang, Zilong and Xu, Shilin and Ji, Shunping and Tong, Yunhai and Qi, Lu and Feng, Jiashi and Yang, Ming-Hsuan},
  journal={arXiv preprint arXiv:2501.04001},
  year={2025}
}

@article{imagen,
  title={Photorealistic text-to-image diffusion models with deep language understanding},
  author={Saharia, Chitwan and Chan, William and Saxena, Saurabh and Li, Lala and Whang, Jay and Denton, Emily L and Ghasemipour, Kamyar and Gontijo Lopes, Raphael and Karagol Ayan, Burcu and Salimans, Tim and others},
  journal={NeurIPS},
  year={2022}
}

@inproceedings{ldm,
  title={High-resolution image synthesis with latent diffusion models},
  author={Rombach, Robin and Blattmann, Andreas and Lorenz, Dominik and Esser, Patrick and Ommer, Bj{\"o}rn},
  booktitle={CVPR},
  year={2022}
}

@inproceedings{animatediff,
  title={AnimateDiff: Animate Your Personalized Text-to-Image Diffusion Models without Specific Tuning},
  author={Guo, Yuwei and Yang, Ceyuan and Rao, Anyi and Liang, Zhengyang and Wang, Yaohui and Qiao, Yu and Agrawala, Maneesh and Lin, Dahua and Dai, Bo},
  booktitle={ICLR},
  year={2024}
}

@article{lavie,
  title={Lavie: High-quality video generation with cascaded latent diffusion models},
  author={Wang, Yaohui and Chen, Xinyuan and Ma, Xin and Zhou, Shangchen and Huang, Ziqi and Wang, Yi and Yang, Ceyuan and He, Yinan and Yu, Jiashuo and Yang, Peiqing and others},
  journal={arXiv preprint arXiv:2309.15103},
  year={2023}
}

@article{modelscope,
  title={Modelscope text-to-video technical report},
  author={Wang, Jiuniu and Yuan, Hangjie and Chen, Dayou and Zhang, Yingya and Wang, Xiang and Zhang, Shiwei},
  journal={arXiv preprint arXiv:2308.06571},
  year={2023}
}

@article{videocrafter1,
  title={Videocrafter1: Open diffusion models for high-quality video generation},
  author={Chen, Haoxin and Xia, Menghan and He, Yingqing and Zhang, Yong and Cun, Xiaodong and Yang, Shaoshu and Xing, Jinbo and Liu, Yaofang and Chen, Qifeng and Wang, Xintao and others},
  journal={arXiv preprint arXiv:2310.19512},
  year={2023}
}

@article{videocrafter2,
  title={Videocrafter2: Overcoming data limitations for high-quality video diffusion models},
  author={Chen, Haoxin and Zhang, Yong and Cun, Xiaodong and Xia, Menghan and Wang, Xintao and Weng, Chao and Shan, Ying},
  journal={arXiv preprint arXiv:2401.09047},
  year={2024}
}

@misc{zeroscope,
  title={Zeroscope},
  author={Tim Brooks and Bill Peebles and Connor Holmes and Will DePue, Yufei Guo and Li Jing and David Schnurr and Joe Taylor and Troy Luhman and Eric Luhman and Clarence Ng and Ricky Wang and Aditya Ramesh},
  journal={https://huggingface.co/cerspense/zeroscope_v2_576w},
  year={2023}
}

@article{CogVideoX,
  title={CogVideoX: Text-to-Video Diffusion Models with An Expert Transformer},
  author={Yang, Zhuoyi and Teng, Jiayan and Zheng, Wendi and Ding, Ming and Huang, Shiyu and Xu, Jiazheng and Yang, Yuanming and Hong, Wenyi and Zhang, Xiaohan and Feng, Guanyu and others},
  journal={CoRR},
  year={2024}
}

@article{HunyuanVideo,
  title={Hunyuanvideo: A systematic framework for large video generative models},
  author={Kong, Weijie and Tian, Qi and Zhang, Zijian and Min, Rox and Dai, Zuozhuo and Zhou, Jin and Xiong, Jiangfeng and Li, Xin and Wu, Bo and Zhang, Jianwei and others},
  journal={arXiv preprint arXiv:2412.03603},
  year={2024}
}

@article{Wan,
  title={Wan: Open and advanced large-scale video generative models},
  author={Wang, Ang and Ai, Baole and Wen, Bin and Mao, Chaojie and Xie, Chen-Wei and Chen, Di and Yu, Feiwu and Zhao, Haiming and Yang, Jianxiao and Zeng, Jianyuan and others},
  journal={arXiv preprint arXiv:2503.20314},
  year={2025}
}

@inproceedings{DreamRelation,
  title     = {DreamRelation: Bridging Customization and Relation Generation},
  author    = {Shi, Qingyu and Qi, Lu and Wu, Jianzong and Bai, Jinbin and Wang, Jingbo and Tong, Yunhai and Li, Xiangtai},
  booktitle = {CVPR},
  year      = {2025}
}

@inproceedings{DeT,
  title={Decouple and Track: Benchmarking and Improving Video Diffusion Transformers for Motion Transfer},
  author={Shi, Qingyu and Wu, Jianzong and Bai, Jinbin and Zhang, Jiangning and Qi, Lu and Li, Xiangtai and Tong, Yunhai},
  booktitle={ICCV},
  year={2025}
}

@article{muddit,
  title={Muddit: Liberating generation beyond text-to-image with a unified discrete diffusion model},
  author={Shi, Qingyu and Bai, Jinbin and Zhao, Zhuoran and Chai, Wenhao and Yu, Kaidong and Wu, Jianzong and Song, Shuangyong and Tong, Yunhai and Li, Xiangtai and Li, Xuelong and others},
  journal={arXiv preprint arXiv:2505.23606},
  year={2025}
}

@article{TangoFlux,
  title={Tangoflux: Super fast and faithful text to audio generation with flow matching and clap-ranked preference optimization},
  author={Hung, Chia-Yu and Majumder, Navonil and Kong, Zhifeng and Mehrish, Ambuj and Bagherzadeh, Amir Ali and Li, Chuan and Valle, Rafael and Catanzaro, Bryan and Poria, Soujanya},
  journal={arXiv preprint arXiv:2412.21037},
  year={2024}
}

@article{Veo3,
  title={Video models are zero-shot learners and reasoners},
  author={Wiedemer, Thadd{\"a}us and Li, Yuxuan and Vicol, Paul and Gu, Shixiang Shane and Matarese, Nick and Swersky, Kevin and Kim, Been and Jaini, Priyank and Geirhos, Robert},
  journal={arXiv preprint arXiv:2509.20328},
  year={2025}
}

@misc{Sora2,
  title = {Sora 2},
  author = {{OpenAI}},
  year = {2025},
  url = {\url{https://openai.com/index/sora-2-system-card/}},
}

@article{SVG,
  title={A simple but strong baseline for sounding video generation: Effective adaptation of audio and video diffusion models for joint generation},
  author={Ishii, Masato and Hayakawa, Akio and Shibuya, Takashi and Mitsufuji, Yuki},
  journal={arXiv preprint arXiv:2409.17550},
  year={2024}
}

@inproceedings{Animate-Sound-Image,
  title={Animate and Sound an Image},
  author={Wang, Xihua and Song, Ruihua and Li, Chongxuan and Cheng, Xin and Li, Boyuan and Wu, Yihan and Wang, Yuyue and Xu, Hongteng and Wang, Yunfeng},
  booktitle={CVPR},
  year={2025}
}

@article{JavisDiT,
  title={Javisdit: Joint audio-video diffusion transformer with hierarchical spatio-temporal prior synchronization},
  author={Liu, Kai and Li, Wei and Chen, Lai and Wu, Shengqiong and Zheng, Yanhao and Ji, Jiayi and Zhou, Fan and Jiang, Rongxin and Luo, Jiebo and Fei, Hao and others},
  journal={arXiv preprint arXiv:2503.23377},
  year={2025}
}

@article{UniVerse-1,
  title={UniVerse-1: Unified Audio-Video Generation via Stitching of Experts},
  author={Wang, Duomin and Zuo, Wei and Li, Aojie and Chen, Ling-Hao and Liao, Xinyao and Zhou, Deyu and Yin, Zixin and Dai, Xili and Jiang, Daxin and Yu, Gang},
  journal={arXiv preprint arXiv:2509.06155},
  year={2025}
}

@article{Bagel,
  title={Emerging properties in unified multimodal pretraining},
  author={Deng, Chaorui and Zhu, Deyao and Li, Kunchang and Gou, Chenhui and Li, Feng and Wang, Zeyu and Zhong, Shu and Yu, Weihao and Nie, Xiaonan and Song, Ziang and others},
  journal={arXiv preprint arXiv:2505.14683},
  year={2025}
}

@article{Show-o,
  title={Show-o: One Single Transformer to Unify Multimodal Understanding and Generation},
  author={Xie, Jinheng and Mao, Weijia and Bai, Zechen and Zhang, David Junhao and Wang, Weihao and Lin, Kevin Qinghong and Gu, Yuchao and Chen, Zhijie and Yang, Zhenheng and Shou, Mike Zheng},
  journal={arXiv preprint arXiv:2408.12528},
  year={2024}
}

@article{Show-o2,
  title={Show-o2: Improved Native Unified Multimodal Models},
  author={Xie, Jinheng and Yang, Zhenheng and Shou, Mike Zheng},
  journal={arXiv preprint arXiv:2506.15564},
  year={2025}
}

@article{Janus-Pro,
  title={Janus-pro: Unified multimodal understanding and generation with data and model scaling},
  author={Chen, Xiaokang and Wu, Zhiyu and Liu, Xingchao and Pan, Zizheng and Liu, Wen and Xie, Zhenda and Yu, Xingkai and Ruan, Chong},
  journal={arXiv preprint arXiv:2501.17811},
  year={2025}
}

@article{Qwen-Image,
  title={Qwen-image technical report},
  author={Wu, Chenfei and Li, Jiahao and Zhou, Jingren and Lin, Junyang and Gao, Kaiyuan and Yan, Kun and Yin, Sheng-ming and Bai, Shuai and Xu, Xiao and Chen, Yilei and others},
  journal={arXiv preprint arXiv:2508.02324},
  year={2025}
}

@article{Emu,
  title={Emu: Generative pretraining in multimodality},
  author={Sun, Quan and Yu, Qiying and Cui, Yufeng and Zhang, Fan and Zhang, Xiaosong and Wang, Yueze and Gao, Hongcheng and Liu, Jingjing and Huang, Tiejun and Wang, Xinlong},
  journal={arXiv preprint arXiv:2307.05222},
  year={2023}
}

@inproceedings{Emu2,
  title={Generative multimodal models are in-context learners},
  author={Sun, Quan and Cui, Yufeng and Zhang, Xiaosong and Zhang, Fan and Yu, Qiying and Wang, Yueze and Rao, Yongming and Liu, Jingjing and Huang, Tiejun and Wang, Xinlong},
  booktitle={CVPR},
  year={2024}
}

@article{Emu3.5,
  title={Emu3. 5: Native Multimodal Models are World Learners},
  author={Cui, Yufeng and Chen, Honghao and Deng, Haoge and Huang, Xu and Li, Xinghang and Liu, Jirong and Liu, Yang and Luo, Zhuoyan and Wang, Jinsheng and Wang, Wenxuan and others},
  journal={arXiv preprint arXiv:2510.26583},
  year={2025}
}

@article{Ming-Omni,
  title={Ming-Omni: A Unified Multimodal Model for Perception and Generation},
  author={AI, Inclusion and Gong, Biao and Zou, Cheng and Zheng, Chuanyang and Zhou, Chunluan and Yan, Canxiang and Jin, Chunxiang and Shen, Chunjie and Zheng, Dandan and Wang, Fudong and others},
  journal={arXiv preprint arXiv:2506.09344},
  year={2025}
}

@article{Ming-Flash-Omni,
  title={Ming-Flash-Omni: A Sparse, Unified Architecture for Multimodal Perception and Generation},
  author={AI, Inclusion and Ma, Bowen and Zou, Cheng and Yan, Canxiang and Jin, Chunxiang and Shen, Chunjie and Zheng, Dandan and Wang, Fudong and Xu, Furong and Yao, GuangMing and others},
  journal={arXiv preprint arXiv:2510.24821},
  year={2025}
}

@article{Qwen2.5-Omni,
  title={Qwen2. 5-omni technical report},
  author={Xu, Jin and Guo, Zhifang and He, Jinzheng and Hu, Hangrui and He, Ting and Bai, Shuai and Chen, Keqin and Wang, Jialin and Fan, Yang and Dang, Kai and others},
  journal={arXiv preprint arXiv:2503.20215},
  year={2025}
}

@inproceedings{CGG,
  title={Betrayed by captions: Joint caption grounding and generation for open vocabulary instance segmentation},
  author={Wu, Jianzong and Li, Xiangtai and Ding, Henghui and Li, Xia and Cheng, Guangliang and Tong, Yunhai and Loy, Chen Change},
  booktitle={ICCV},
  year={2023}
}

@inproceedings{LGVI,
  title={Towards language-driven video inpainting via multimodal large language models},
  author={Wu, Jianzong and Li, Xiangtai and Si, Chenyang and Zhou, Shangchen and Yang, Jingkang and Zhang, Jiangning and Li, Yining and Chen, Kai and Tong, Yunhai and Liu, Ziwei and others},
  booktitle={CVPR},
  year={2024}
}

@inproceedings{DiffSensei,
  title={Diffsensei: Bridging multi-modal llms and diffusion models for customized manga generation},
  author={Wu, Jianzong and Tang, Chao and Wang, Jingbo and Zeng, Yanhong and Li, Xiangtai and Tong, Yunhai},
  booktitle={CVPR},
  year={2025}
}

@article{VideoJAM,
  title={Videojam: Joint appearance-motion representations for enhanced motion generation in video models},
  author={Chefer, Hila and Singer, Uriel and Zohar, Amit and Kirstain, Yuval and Polyak, Adam and Taigman, Yaniv and Wolf, Lior and Sheynin, Shelly},
  journal={arXiv preprint arXiv:2502.02492},
  year={2025}
}

@inproceedings{UDPDiff,
  title={Unified Dense Prediction of Video Diffusion},
  author={Yang, Lehan and Qi, Lu and Li, Xiangtai and Li, Sheng and Jampani, Varun and Yang, Ming-Hsuan},
  booktitle={CVPR},
  year={2025}
}

@inproceedings{VGGSound,
  title={Vggsound: A large-scale audio-visual dataset},
  author={Chen, Honglie and Xie, Weidi and Vedaldi, Andrea and Zisserman, Andrew},
  booktitle={ICASSP},
  year={2020}
}

@inproceedings{AVSync15,
  title={Audio-synchronized visual animation},
  author={Zhang, Lin and Mo, Shentong and Zhang, Yijing and Morgado, Pedro},
  booktitle={ECCV},
  year={2024},
}

@inproceedings{Landscape,
  title={Sound-guided semantic video generation},
  author={Lee, Seung Hyun and Oh, Gyeongrok and Byeon, Wonmin and Kim, Chanyoung and Ryoo, Won Jeong and Yoon, Sang Ho and Cho, Hyunjun and Bae, Jihyun and Kim, Jinkyu and Kim, Sangpil},
  booktitle={ECCV},
  year={2022},
}

@inproceedings{TheGreatestHits,
  title={Visually indicated sounds},
  author={Owens, Andrew and Isola, Phillip and McDermott, Josh and Torralba, Antonio and Adelson, Edward H and Freeman, William T},
  booktitle={CVPR},
  year={2016}
}

@article{AdamW,
  title={Decoupled weight decay regularization},
  author={Loshchilov, Ilya and Hutter, Frank},
  journal={arXiv preprint arXiv:1711.05101},
  year={2017}
}

@inproceedings{VBench,
  title={Vbench: Comprehensive benchmark suite for video generative models},
  author={Huang, Ziqi and He, Yinan and Yu, Jiashuo and Zhang, Fan and Si, Chenyang and Jiang, Yuming and Zhang, Yuanhan and Wu, Tianxing and Jin, Qingyang and Chanpaisit, Nattapol and others},
  booktitle={CVPR},
  year={2024}
}

@article{Videophy-2,
  title={Videophy-2: A challenging action-centric physical commonsense evaluation in video generation},
  author={Bansal, Hritik and Peng, Clark and Bitton, Yonatan and Goldenberg, Roman and Grover, Aditya and Chang, Kai-Wei},
  journal={arXiv preprint arXiv:2503.06800},
  year={2025}
}

@inproceedings{CLAP,
  title={Clap learning audio concepts from natural language supervision},
  author={Elizalde, Benjamin and Deshmukh, Soham and Al Ismail, Mahmoud and Wang, Huaming},
  booktitle={ICASSP},
  year={2023},
}

@article{FAD,
  title={Fr$\backslash$'echet audio distance: A metric for evaluating music enhancement algorithms},
  author={Kilgour, Kevin and Zuluaga, Mauricio and Roblek, Dominik and Sharifi, Matthew},
  journal={arXiv preprint arXiv:1812.08466},
  year={2018}
}

@inproceedings{Synchformer,
  title={Synchformer: Efficient synchronization from sparse cues},
  author={Iashin, Vladimir and Xie, Weidi and Rahtu, Esa and Zisserman, Andrew},
  booktitle={ICASSP},
  year={2024},
}
}

\appendix
\twocolumn[{
\centering
\section*{Does Hearing Help Seeing? Investigating Audio–Video Joint Denoising \\ for Video Generation Supplementary Material}
\vspace{5mm} 
}]

\noindent
\textbf{Overview.}
\begin{itemize}
    \item \textbf{\cref{sec:supp-introduction-video}}: Introduction video.
    \item \textbf{\cref{sec:supp-more-qualitative-results}}: More qualitative results.
    \item \textbf{\cref{sec:more-implementation-details}}: More implementation details.
    \item \textbf{\cref{sec:supp-user-study}}: User study between T2AV and T2V results.
    \item \textbf{\cref{sec:supp-more-ablation-study}}: More ablation studies.
    \item \textbf{\cref{sec:supp-limitations}}: Limitations and failure cases.
    \item \textbf{\cref{sec:supp-broader-impacts}}: Broader impacts.
    \item \textbf{\cref{sec:supp-vggsound-subset-details}}: VGGSound subset details.
\end{itemize}

\section{Introduction Video}
\label{sec:supp-introduction-video}
To help readers quickly grasp the primary idea of our work, we provide a 6-minute introduction video. Please refer to ``\textbf{introduction\_video.mp4}'' in the supplementary file.

\section{More Qualitative Results}
\label{sec:supp-more-qualitative-results}
Due to the large amount of qualitative results, we chose to present them in a separate, anonymous local HTML page. Please refer to ``\textbf{demo/index.html}'' in the supplementary file. On this project page, we provide extensive qualitative results, making a comprehensive comparison between T2AV results (with audio) and T2V results.

\section{More Implementation Details}
\label{sec:more-implementation-details}
The audio and video loss weights are set to 1:1 in all our experiments, except for ablation experiments.
For all generations, we use the same random seed for T2AV and T2V. We observe that their results from the same prompts share similar visual layouts.

\begin{figure}[!ht]
	\centering
	\includegraphics[width=1.0\linewidth]{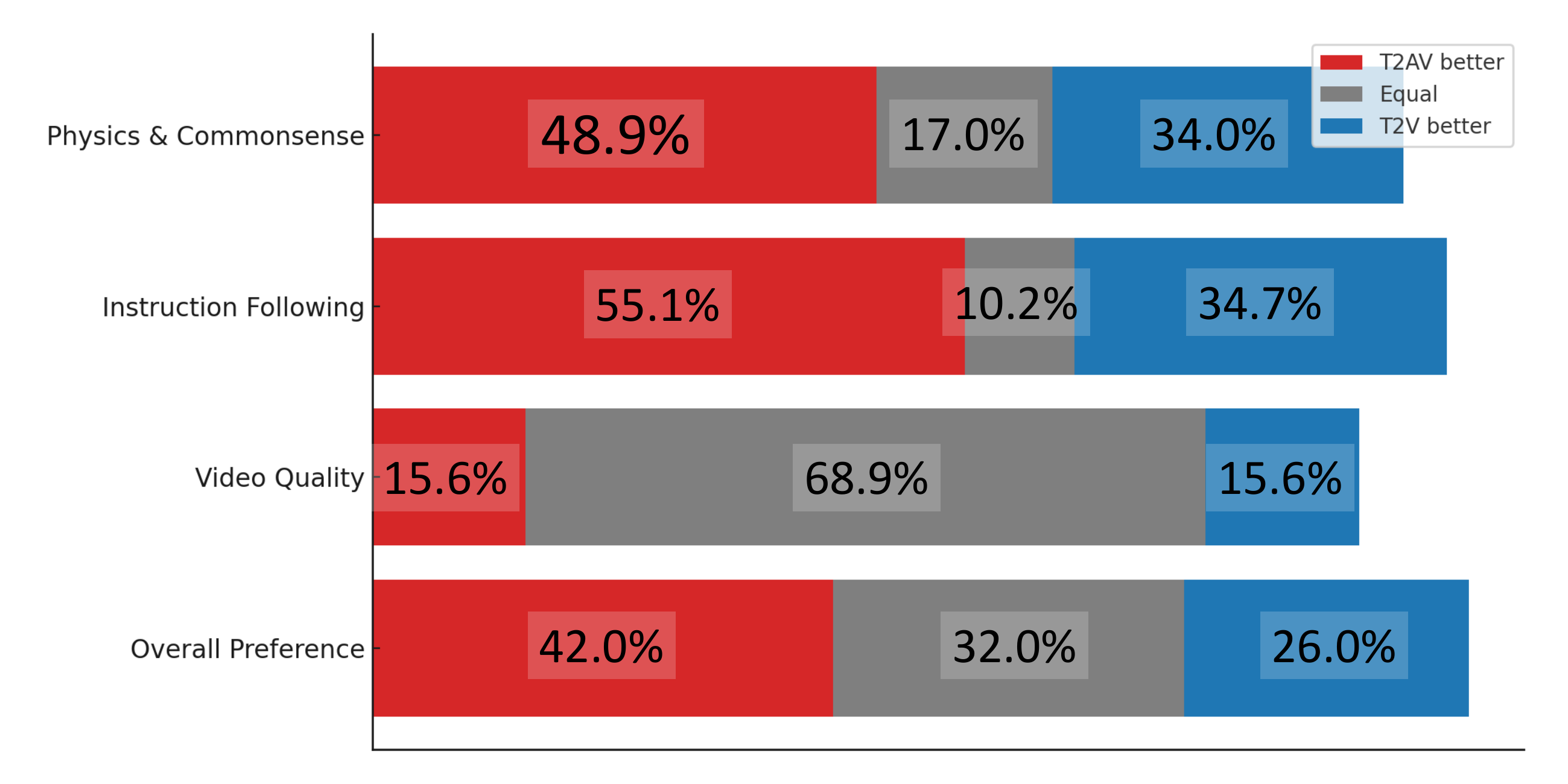}
	\caption{\textbf{User Study.} In all aspects of evaluation, T2AV is generally better than T2V.}
	\label{fig:supp-user-study}
\end{figure}

\section{User Study}
\label{sec:supp-user-study}
To provide a comprehensive qualitative assessment of video generation performance, we conduct a subjective user study comparing T2AV against the T2V baseline. We randomly sampled 50 generation pairs from the ALT-Merge and VGGSound evaluation sets. Participants are asked to evaluate the videos based on four distinct criteria: Overall Preference, a general impression of the video without restriction to specific dimensions. Video Quality, the visual fidelity and clarity of the imaging. Instruction Following, the degree of alignment between the generated content and the video prompt. Physics \& Commonsense, the plausibility of physical dynamics and interactions within the scene. To strictly isolate the visual improvements gained from joint training and exclude auditory bias, we remove the generated audio tracks from the T2AV outputs. Consequently, both T2AV and T2V samples were presented as silent videos. The presentation order is randomized and blinded to prevent evaluators from identifying the source model.
We recruit a diverse panel of 10 evaluators varying in gender, age, and occupation; the majority had no professional background in AI, ensuring the feedback reflects a general user perspective. Results are aggregated by averaging responses across all participants. The results are summarized in~\cref{fig:supp-user-study}. The T2AV model matches or outperforms the T2V baseline across all four dimensions. Notably, T2AV demonstrates a significant advantage in Overall Preference, Instruction Following, and Physics \& Commonsense. This superiority in physical plausibility and prompt adherence strongly supports our hypothesis that audio serves as a strong regularizer for learning world dynamics. These subjective findings corroborate our quantitative metrics, confirming that hearing indeed helps seeing in video generation.

\section{More Ablation Studies}
\label{sec:supp-more-ablation-study}

\begin{table}[!ht]
    \centering
    \caption{\textbf{Ablation on adjusting audio and video loss weights.} $\lambda_a$ and $\lambda_v$ are audio and video loss weights, respectively. FAD is in units of $10^4$. All ablations are done on ALT-Merge.}
    \label{tab:supp-ablation-loss-weight}
    \resizebox{1.0\linewidth}{!}{
    \begin{tabular}{l|cccccc}
        \toprule
        Loss Weights & FAD $\downarrow$ & CLAP $\uparrow$ & IQ $\uparrow$ & SC $\uparrow$ & PH $\uparrow$ & Sync $\downarrow$ \\
        \midrule
        $\lambda_a = 0.1, \lambda_v = 1$ & 10.01 & 35.09 & 56.11 & 95.05 & \textbf{4.37} & 0.34 \\
        $\lambda_a = 0.3, \lambda_v = 1$ & 9.95 & 36.56 & 58.87 & 95.71 & 4.36 & 0.35 \\
        \textbf{$\lambda_a = 1, \lambda_v = 1$ (Ours)} & \textbf{9.36} & \textbf{38.75} & \textbf{59.87} & \textbf{96.30} & 4.36 & \textbf{0.29} \\
        \bottomrule
    \end{tabular}}
\end{table}

\noindent
\textbf{Audio and video loss weight.}
To further enhance video learning, we investigate adjusting the relative loss weight. We test audio loss weights $\lambda_a$ of 0.3 and 0.1, increasing the proportion of video loss in the total loss. The results in~\cref{tab:supp-ablation-loss-weight} indicate that reducing the audio loss weight leads to a significant degradation in audio-related metrics, including FAD and CLAP. Conversely, video quality metrics such as IQ and SC do not show meaningful improvement and are highest when the weights are balanced. This finding aligns with our conclusion that incorporating audio in training aids video quality enhancement.
Notably, the Physics metric experiences a marginal increase of 0.01 at an audio weight of 0.1. Still, it remains unchanged at 0.3, suggesting that the relative loss weight has a minimal impact on the model's understanding of physical principles. The crucial factor appears to be the presence of audio information itself. Furthermore, the Synchronization metric deteriorates substantially as the audio loss weight is reduced, indicating that balanced loss weights are essential for achieving optimal audio-visual synchronization.

\begin{table}[!ht]
    \centering
    \caption{\textbf{Ablation on audio guidance scale in inference.} }
    \label{tab:supp-ablation-audio-guidance-scale}
    \resizebox{1.0\linewidth}{!}{
    \begin{tabular}{c|cccccc}
        \toprule
        Scale & FAD $\downarrow$ & CLAP $\uparrow$ & IQ $\uparrow$ & SC $\uparrow$ & PH $\uparrow$ & Sync $\downarrow$ \\
        \midrule
        4.0 & 9.47 & 38.54 & 59.86 & \textbf{96.31} & 4.35 & \textbf{0.29} \\
        \textbf{4.5 (Ours)} & \textbf{9.36} & 38.75 & \textbf{59.87} & 96.30 & \textbf{4.36} & \textbf{0.29} \\
        5.0 & 9.55 & \textbf{39.67} & 59.69 & \textbf{96.31} & 4.33 & \textbf{0.29} \\
        \bottomrule
    \end{tabular}}
\end{table}

\noindent
\textbf{Guidance scale.}
We conduct an ablation study on the audio guidance scale for classifier-free guidance during inference. As shown in~\cref{tab:supp-ablation-audio-guidance-scale}, varying the scale from 4.0 to 5.0 leads to only marginal fluctuations across all evaluation metrics. This finding highlights the stability of our model's performance with respect to this inference hyperparameter. We ultimately adopt a scale of 4.5, which aligns with the default setting of the pre-trained TangoFlux model~\cite{TangoFlux}.

\begin{figure}[!ht]
	\centering
	\includegraphics[width=1.0\linewidth]{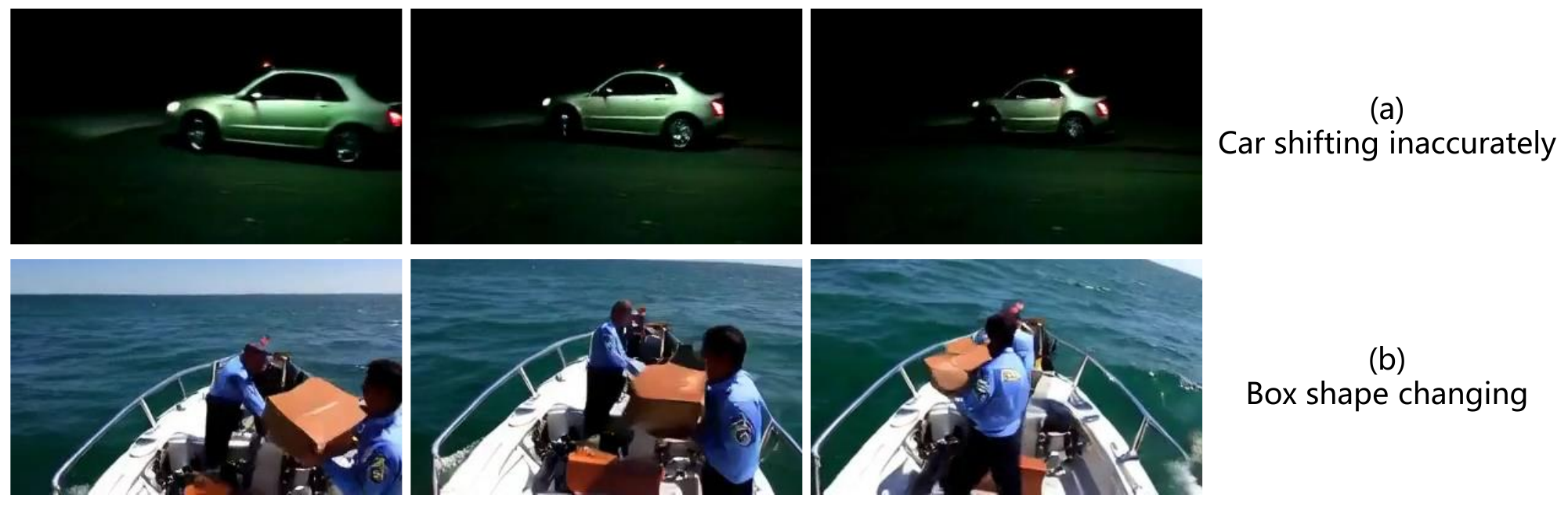}
	\caption{\textbf{Failure cases.} (a) The car drifts to the side, instead of forward or backward. (b) The shape of the box changes.}
	\label{fig:supp-limitation}
\end{figure}

\section{Limitations}
\label{sec:supp-limitations}
While our work presents compelling evidence for the benefits of audio-video joint denoising, we acknowledge several limitations that offer avenues for future research.
First, a notable limitation of our current T2AV model is its inability to generate clear and coherent human speech. While it can produce sounds indicative of human vocalization, the output lacks articulatory clarity. This is primarily due to the nature of our training data and annotations; the audio captions describe acoustic events (e.g., "a woman speaking") but do not contain verbatim transcriptions of spoken content. Consequently, the model learns to associate the visual of a person talking with the general sound of speech, rather than specific words.
Interestingly, despite this, we observe that the model sometimes generates muffled speech containing words semantically related to the visual context. This suggests that the model has begun to build a rudimentary association between specific scenes and relevant vocabulary, forming a partial world model of spoken language.
Second, although our T2AV model demonstrates improved physical commonsense, it is not immune to generating content that occasionally violates physical laws. It still produces failure cases, as shown in~\cref{fig:supp-limitation}. This is a recognized and persistent challenge across the broader field of video generation. 
However, it is crucial to emphasize that these instances of physical implausibility are less frequent and less severe than in the T2V-only counterpart, as supported by our quantitative and qualitative results. Therefore, this limitation does not undermine our central finding that incorporating an audio objective serves as a strong regularizer, leading to a significant overall improvement in physical realism compared to a video-only baseline.

\section{Broader Impacts}
\label{sec:supp-broader-impacts}
Our research demonstrates that incorporating an audio generation objective significantly improves the quality of video generation, particularly in depicting physically plausible dynamics and interactions. This finding has two main broader implications.

\noindent
\textbf{Informing the development of world models.}
The central finding—that ``hearing helps seeing'' suggests that building more comprehensive world models may depend on integrating a wider range of sensory inputs. Our work indicates that audio can serve as a privileged signal, providing crucial information about causality and physical interactions (e.g., collisions that produce sound) that is less ambiguous than visual data alone. This provides a compelling rationale for incorporating audio and other modalities into the development of more robust and physically grounded AI systems that can better comprehend and reason about the world.

\noindent
\textbf{Advancing unified multimodal architectures.}
This study contributes to the growing body of research on unified multimodal models. By demonstrating that one modality (audio) can serve as a regularizer to enhance the quality of generation of another (video), our work supports the hypothesis that different data streams can provide complementary supervisory signals. This insight can guide the development of more effective and efficient unified architectures, enabling knowledge transfer across modalities to achieve a more holistic understanding and generative capability.

\section{VGGSound Subset Details}
\label{sec:supp-vggsound-subset-details}
As outlined in our experimental setup, we partition the VGGSound dataset into two distinct subsets, ``AV-Tight'' and ``AV-Loose,'' to better evaluate our model's handling of different degrees of audio-visual correspondence. This division, detailed in~\cref{tab:supp-vggsound-subsets}, is based on the causal relationship between the visual events and their corresponding sounds.
The AV-Tight subset includes categories where the audio is directly and immediately linked to the visible action. In these instances, there is a strong, predictable correlation between what is seen and what is heard. For example, classes such as chopping\_wood, typing\_on\_computer\_keyboard, and striking\_pool feature sounds that are intrinsically linked to and caused by specific, observable physical interactions.
In contrast, the AV-Loose subset contains categories where the relationship between the audio and video is more ambient or contextual, rather than strictly causal and synchronized. While the sound and visuals co-occur, the sound is not necessarily produced by a single, discrete visual event. For instance, in categories like airplane\_flyby, raining, or wind\_noise, the audio represents the general acoustic environment rather than a sound tied to a specific, synchronized action. This distinction allows us to test the hypothesis that joint audio-video denoising provides the most significant benefits when the modalities are tightly coupled and mutually informative.

\begin{table*}[!t]
\centering
\caption{List of AV-Tight and AV-Loose subsets in VGGSound~\cite{VGGSound}.}
\label{tab:supp-vggsound-subsets} 
\scalebox{0.8}{ 
\begin{tabular}{lll}
\toprule
\multicolumn{3}{c}{\textbf{AV-Tight Classes}} \\
\midrule
arc\_welking & bathroom\_ventilation\_fan\_running & beat\_boxing \\
cattle,\_bovinae\_cowbell & chainsawing\_trees & child\_speech,\_kid\_speaking \\
chopping\_food & chopping\_wood & church\_bell\_ringing \\
cutting\_hair\_with\_electric\_trimmers & disc\_scratching & driving\_motorcycle \\
eletric\_grinder\_grinding & eletric\_shaver,\_electric\_razor\_shaving & eletric\_blender\_running \\
firing\_cannon & firing\_muskets & forging\_swords \\
metronome & missile\_launch & motorboat,\_speedboat\_acceleration \\
ocean\_burbling & opening\_or\_closing\_car\_eletric\_windows & planing\_timber \\
playing\_guiro & playing\_saxophone & printer\_printing \\
ripping\_paper & rope\_skipping & sharpen\_knife \\
singing\_bowl & skateboarding & spraying\_water \\
strike\_lighter & striking\_pool & toilet\_flushing \\
typing\_on\_computer\_keyboard &  &  \\
\midrule[\heavyrulewidth] 
\multicolumn{3}{c}{\textbf{AV-Loose Classes}} \\
\midrule
air\_conditioning\_noise & air\_horn & airplane \\
airplane\_flyby & alligators,\_crocodiles\_hissing & ambulance\_siren \\
baby\_crying & baby\_laughter & baltimore\_oriole\_calling \\
barn\_swallow\_calling & basketball\_bounce & bee,\_wasp,\_etc,\_buzzing \\
bird\_chirping,\_tweeting & bird\_squawking & blowtorch\_igniting \\
bouncing\_on\_trampoline & bowling\_impact & bull\_bellowing \\
canary\_calling & cap\_gun\_shooting & car\_engine\_idling \\
car\_engine\_starting & car\_passing\_by & cat\_caterwauling \\
cat\_mewwing & cat\_purring & cattle\_mooing \\
cell\_phone\_buzzing & cheetah\_chirrup & chicken\_clucking \\
chicken\_crowing & child\_singing & chimpanzee\_pant-hooting \\
chinchilla\_barking & chipmunk\_chirping & civil\_defense\_siren \\
cow\_lowing & coyote\_howling & cricket\_chirping \\
crow\_cawing & cuckoo\_bird\_calling & cupboard\_opening\_or\_closing \\
dinosaurs\_bellowing & dog\_barking & dog\_bow-bow \\
dog\_growling & dog\_howling & donkey,\_ass\_braying \\
driving\_buses & driving\_snowmobile & duck\_quacking \\
eagle\_screaming & eating\_with\_cutlery & elephant\_trumpeting \\
elk\_bugling & engine\_accelerating,\_revving,\_vroom & female\_singing \\
female\_speech,\_woman\_speaking & ferret\_dooking & fire\_truck\_siren \\
fly\_housefly\_buzzing & foghorn & footsteps\_on\_snow \\
francolin\_calling & frog\_croaking & gibbon\_howling \\
man\_speech,\_man\_speaking & mosquito\_buzzing & mouse\_pattering \\
mouse\_squeaking & mynah\_bird\_singing & opening\_or\_closing\_drawers \\
orchestra & otter\_growling & owl\_hooting \\
parrot\_talking & penguins\_braying & people\_burping \\
people\_hiccup & people\_running & prople\_screaming \\
people\_sneezing & pheasant\_crowing & pig\_oinking \\
pigeon,\_dove\_cooing & plastic\_bottle\_crushing & playing\_flute \\
playing\_harpsichord & playing\_washboard & police\_car(siren) \\
police\_radio\_chatter & popping\_popcorn & pumping\_water \\
race\_car,\_auto\_racing & railroad\_car,\_train\_wagon & raining \\
rapping & reversing\_beeps & roller\_coaster\_running \\
rowboat,\_canoe,-kayak\_rowing & running\_electric\_fan & sailing \\
scuba\_diving & sea\_lion\_barking & sea\_waves \\
sheep\_bleating & shot\_football & singing\_choir \\
skidding & skiing & sliding\_door \\
sloshing\_water & slot\_machine & smoke\_detector\_beeping \\
snake\_hissing & snake\_rattling & splashing\_water \\
squishing\_water & stream\_burbling & striking\_bowling \\
subway,\_metro,\_underground & swimming & tap\_dancing \\
tapping\_guitar & telephone\_bell\_ringing & thunder \\
tornado\_roaring & tractor\_digging & train\_horning \\
train\_wheels\_squealing & train\_whistling & turkey\_gobbling \\
underwater\_bubbling & using\_sewing\_machines & vacuum\_cleaner\_cleaning\_floors \\
vehicle\_horn,\_car\_horn,\_honking & volcano\_explosion & warbler\_chirping \\
waterfall\_burbling & whale\_calling & wind\_noise \\
wood\_thrush\_calling & woodpecker\_pecking\_tree & writing\_on\_blackboard\_with\_chalk \\
yodelling & zebra\_braying &  \\
\bottomrule
\end{tabular}}
\end{table*}





\end{document}